\pdfoutput=1
\documentclass[11pt]{article}

\usepackage{coling}
\usepackage{times}
\usepackage{latexsym}
\usepackage[T1]{fontenc}
\usepackage[utf8]{inputenc}
\usepackage{microtype}
\usepackage{inconsolata}
\usepackage{graphicx}
\usepackage[most]{tcolorbox}
\usepackage{url}
\usepackage{multirow}
\usepackage{booktabs}
\usepackage{colortbl}
\usepackage{amssymb}
\usepackage{subfigure}
\usepackage{tcolorbox}
\usepackage{listings}
\usepackage{xcolor}

\definecolor{rq1color}{RGB}{22, 160, 133}   % Teal
\definecolor{rq2color}{RGB}{41, 128, 185}   % Blue
\definecolor{rq3color}{RGB}{142, 68, 173}   % Amethyst
\definecolor{rq4color}{RGB}{230, 126, 34}   % Carrot Orange
\usepackage{subcaption} 
\usepackage{booktabs}
\usepackage{pdfpages}
\usepackage[table,xcdraw]{xcolor}
\usepackage{placeins}
\usepackage{tabularx}    % X column (auto wrap)
\usepackage{array}       % m{width} column type
\usepackage{ragged2e}    % \RaggedRight for X columns
\usepackage{enumitem}
\usepackage{todonotes}
\newcolumntype{Y}{>{\RaggedRight\arraybackslash}X} % ragged X

\newcommand{\datasetname}{Afri-MCQA} %
\title{\datasetname: Multimodal Cultural Question Answering \\ for African Languages}

\author{\normalsize Atnafu Lambebo Tonja$^{1,*}$, Srija Anand$^{1,2,*}$, Emilio Villa-Cueva$^{1}$ \thanks{\ \ \ Equal contribution}, Israel Abebe Azime$^{3}$, \\
\textbf{\normalsize Jesujoba O. Alabi$^{3}$, Muhidin A. Mohamed$^{4}$, Debela Desalegn Yadeta$^{5}$, Negasi Haile Abadi$^{6}$, } \\
\textbf{\normalsize Abigail Oppong$^{7}$, Nnaemeka Casmir Obiefuna$^{8}$, Idris Abdulmumin$^{9}$, Naome A. Etori$^{10}$,  }\\
\textbf{\normalsize Eric Peter Wairagala$^{11}$, Kanda Patrick Tshinu$^{12}$, Imanigirimbabazi Emmanuel$^{13}$, }\\
\textbf{\normalsize Gabofetswe Malema$^{14}$, Alham Fikri Aji$^{1}$, David Ifeoluwa Adelani$^{15}$, Thamar Solorio$^{1}$ }\\\\
\footnotesize
$^1$MBZUAI, $^2$AI4Bharat, Indian Institute of Technology, Madras, $^3$Saarland University,  $^{4}$Aston University,\\ \footnotesize 
$^5$Addis Ababa University, $^6$Lesan AI, $^7$Independent, $^{8}$Friedrich-Alexander University, $^9$University of Pretoria,\\
\footnotesize 
  $^{10}$University of Minneosta -Twin Cities, $^{11}$Lelapa AI,$^{12}$Tshwane University of Technology, \\
 \footnotesize
 $^{13}$Digital Umuganda, $^{14}$University of Botswana, $^{15}$Mila, McGill University \& Canada CIFAR AI Chair \\
}

\begin{document}
\maketitle
\begin{abstract}
Africa is home to over one-third of the world's languages, yet remains underrepresented in AI research. We introduce \datasetname, the first \textbf{M}ultilingual \textbf{C}ultural \textbf{Q}uestion-\textbf{A}nswering benchmark covering 7.5k Q\&A pairs across 15 African languages from 12 countries. The benchmark offers parallel English-African language Q\&A pairs across text and speech modalities and was entirely created by native speakers. Benchmarking large language models (LLMs) on \datasetname~ shows that  open-weight models show poor performance across evaluated cultures, with near-zero accuracy on open-ended VQA when queried using native language or speech. 
To evaluate linguistic competence, we include control experiments meant to assess this specific aspect separate from cultural knowledge, and we observe significant performance gaps between native languages and English for both text and speech. These findings underscore the need for speech-first approaches, culturally grounded pretraining, and cross-lingual cultural transfer. To support more inclusive multimodal AI development in African languages, we release our \datasetname~ under academic license or CC BY-NC 4.0 on HuggingFace\footnote{https://huggingface.co/datasets/Atnafu/Afri-MCQA}.

\end{abstract}

\section{Introduction}
%     \item Multi-modal LLM and Evaluation
%     \item African languages and limitations

%\url{https://arxiv.org/pdf/2203.08351}

% \textcolor{blue}{

%\paragraph{Focus on Africa} 
Africa is one of the most culturally diverse and rapidly growing regions in the world. It is home to more than one-third of the world's languages \cite{hammarstrom2018survey} and a population exceeding 1.3 billion, projected to surpass 2.5 billion by 2050 \cite{Simane2025}. African languages have linguistic features that are very different from many high-resource languages represented in LLMs, including rich morphology, use of noun classes, tonality, and serial verb constructions, among others~\citep{Van_de_Velde2006-fz,adebara-abdul-mageed-2022-towards}. %Linguistic diversity in the region includes features such as serial verb constructions, vowel harmony, and tone \cite{adebara-abdul-mageed-2022-towards}, which differ from many high-resource languages around which NLP has been developed. 
Additionally, many African languages are primarily spoken, with literacy often occurring in a colonial or foreign language. This makes speech-based applications such as Multimodal large language models (MLLMs) particularly relevant for enabling access to technology.
% % }
\begin{figure}[t]
    \centering
    \includegraphics[width=\linewidth]{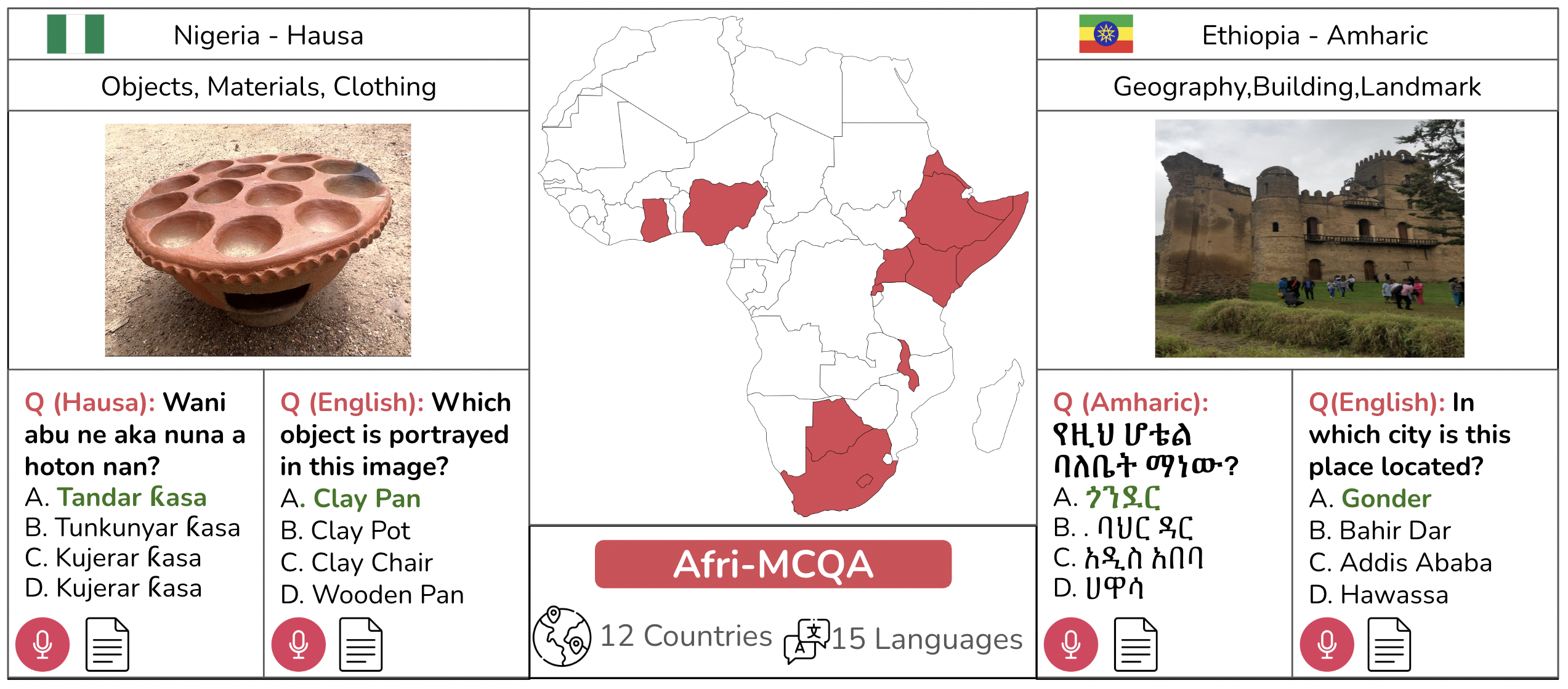}
    %\caption{Afri-MCQA, a multilingual and multimodal cultural question answering benchmark.} 
    \caption{Examples of \datasetname~ datapoints, containing parallel text and speech QA pairs grounded in culturally relevant images across English and native African languages.} 
    \label{fig:dataset_exam}
    \vspace{-4mm}
\end{figure}

%% need to connect better with paragraph below
%     \item Progress on benchmark creation
% In recent years, researchers \cite{} and grassroots communities such as Masakhane \footnote{https://www.masakhane.io/} \cite{}, EthioNLP \footnote{https://ethionlp.github.io/} \cite{}, Nigeria-hauNLP \footnote{https://Nigeria-haunlp.github.io/} \cite{} have made significant progress in developing evaluation benchmarks for different downstream NLP tasks \cite{}. These resources have facilitated research in many African languages that had been previously overlooked. However, most of these evaluations are designed for text-only downstream tasks, such as news classification \cite{}, named entity recognition \cite{}, sentiment analysis \cite{}, question answering \cite{}, and intent detection \cite{}. Due to the complexity of data collection at scale in these communities, some of the available resources, such as IrokoBench \citet{}, GlobalMMLU \citet{}, have been created by translating English benchmarks. 
MLLMs perform well on tasks that require reasoning over visual, %auditory, 
spoken, and textual inputs to follow user instructions~\cite{liu-etal-2021-visually,gemmateam2025gemma3technicalreport,abdin2024phi4technicalreport}. %Most MLLMs are built on top of a large language model backbone, combined with modality-specific encoders that project inputs into a shared embedding space \cite{liu-etal-2021-visually,gemmateam2025gemma3technicalreport,abdin2024phi4technicalreport}.
However, these systems are developed primarily for high-resource, mostly Western-centric, or as  \citet{mihalcea2024aiweirdwayai} put it, they have a `WEIRD' coverage. As a result, their knowledge and reasoning abilities tend to favor the languages and cultures represented in their training data \cite{mihalcea2024aiweirdwayai}.
 % , despite the coverage of African contexts remaining limited.
 % , especially African contexts with fewer represented languages~\cite{romero2024cvqa, vayani2025all}. %they tend to benefit the languages and cultures represented in their training data. 
%Recent benchmarks have begun to assess global cultural knowledge through question answering \cite{romero2024cvqa, vayani2025all}, despite the coverage of African contexts remaining limited.

Although most evaluation datasets for low-resource languages are translated~\cite{nllb2022,adelani-etal-2024-sib,adelani-etal-2025-irokobench,azime-etal-2024-walia,alabi-etal-2025-charting}, researchers have recently made promising progress in creating culturally relevant benchmarks for African languages across various NLP tasks \cite{adelani-etal-2023-masakhanews,azime-etal-2025-proverbeval,muhammad-etal-2025-afrihate,yu-etal-2025-injongo,tonja2024inkubalm,tonja2024ethiollm}. However, these evaluations are designed for text-only tasks. In the language-vision space, recent benchmarks assess global cultural knowledge through question answering  \cite{romero2024cvqa, vayani2025all, winata-etal-2025-worldcuisines}, yet their coverage of African languages and contexts remains limited.

%some African languages have been included in multilingual and multimodal benchmarks that cover various regions. For example, CVQA \cite{romero2024cvqa} probes visual cultural knowledge and includes three African languages, while the worldwide recipe benchmark \cite{magomere2025world} has broader coverage but focuses only on food categories, including 7.87\% of data from Africa. Hence, evaluation in this space is still limited to a few benchmarks that have not been specifically designed for the continent.

%These resources have facilitated research in many African languages that had been previously overlooked. However, most of these evaluations are designed for text-only downstream tasks. In the multimodal setting, some African languages have been included in multilingual and multimodal benchmarks that cover various regions. For example, CVQA \cite{romero2024cvqa} probes visual cultural knowledge and includes three African languages, while the worldwide recipe benchmark \cite{magomere2025world} has broader coverage but focuses only on food categories, including 7.87\% of data from Africa. Hence, evaluation in this space is still limited to a few benchmarks that have not been specifically designed for the continent.

Africa's rich cultural diversity demands AI systems that can effectively serve its communities. A critical step toward this goal is evaluating how well current MLLMs understand and reason about African cultural knowledge in multimodal settings.
%As noted, existing benchmarks have several limitations: \textbf{(1)} most support only text, \textbf{(2)} some are translations of English benchmarks, and \textbf{(3)} few African languages are represented in current multimodal benchmarks.
% \item our work and contributions 
To enable this, we introduce \datasetname~, \textbf{the first %audiovisual, 
multilingual cultural VQA dataset supporting text and speech modalities} (as shown in \autoref{fig:dataset_exam}). The dataset consists of $\approx7.5k$ Q\&A pairs (in English and native languages) across 15 African languages from 12 countries. Our data collection involves native speakers as annotators, who reside in the countries where their respective languages are spoken. 
%To ensure high-quality data, we included native speakers as coordinators for each language, who are familiar with the culture of the countries involved in our work. 
Each language covers $\approx$ 500 image-grounded Q\&A samples, with both text and spoken audio in the native language and English. We evaluate multiple MLLMs across various setups to answer the following research questions:

\begin{table*}[h!]
\centering
\small
 \resizebox{0.95\linewidth}{!}{%
\begin{tabular}{lrrrrccc}
\hline
 Datasets& \textbf{\# African Lang} & \textbf{\# Countries} & \textbf{QA categories} & \textbf{\# QA per langs} & \textbf{Audio QA} &\textbf{Parallel Data}  \\ \hline
 CVQA \cite{romero2024cvqa}&5&4&10&200&$\times$&$\checkmark$\\
 WC-VQA \cite{winata-etal-2025-worldcuisines}&1&1&1&-&$\times$&$\checkmark$ \\ 
  M5 \cite{schneider2024m5}&4&4&-&-&$\times$&$\times$\\\
  HaVQA \cite{parida2023havqa}&1&1&-&6,200&$\times$&$\checkmark$\\\hline
 \textbf{\datasetname{} (ours)}&15&12&10&500&$\checkmark$&$\checkmark$\\ \hline
\end{tabular}}
\caption{Data statistics for Afri-MCQA compared to existing VQA datasets that include African languages.}
\label{tab:dataset_comparison}
\vspace{-2mm}
\end{table*}
% \colorbox{rq1}{\textbf{RQ1}}
\colorbox{rq1color}{\textbf{\color{white}RQ1\color{black}}}: How well do MLLMs understand African cultural contexts in visually-grounded QA? 

\colorbox{rq2color}{\color{white}\textbf{RQ2}\color{black}}: How does input modality (text vs. speech) affect performance?

\colorbox{rq3color}{\color{white}\textbf{RQ3}\color{black}}: How does query language (native vs. English) affect performance, and do differences reflect language understanding or cultural knowledge gaps? 

\colorbox{rq4color}{\color{white}\textbf{RQ4}\color{black}}: How does task format (Multiple-Choice vs. Open QA) affect accuracy?

% % We outline below the contributions of our work:
% Through this evaluation, we aim to answer the following research questions:
% \begin{itemize}
%     \item  \textcolor{rq1color}{\textbf{RQ1}}: How well do MLLMs \textit{know and understand} African cultural contexts in visually-grounded QA tasks?
%      \item   \textcolor{rq2color}{\textbf{RQ2}}: What is the current performance gap in state-of-the-art MMLMs when queried with text vs. speech in African languages?
%      \item   \textcolor{rq3color}{\textbf{RQ3}}: How does the query language affect the model's response on visual cultural knowledge?
%      % Do models generalize across native vs. English QA?
%     \item  \textcolor{rq4color} {\textbf{RQ4}}: How does accuracy differ when evaluated with Multiple-Choice QA compared to Open QA?
%     % In which modalities (text or speech) do the current SOTA MMLMs perform well for African languages?
%     \item  \textcolor{rq5color}{\textbf{RQ5}}: Do performance differences between native and English QA reflect language understanding or cultural knowledge gaps? 
% \end{itemize}

Our contributions are: \textbf{(1)} We introduce \datasetname, the first large-scale multilingual visual cultural QA benchmark for 15 African languages across 12 countries, with parallel text and speech QA created by native speakers. \textbf{(2)} We demonstrate that text-based multilingual ability does not transfer to speech understanding, emphasizing the need for more African language representation in multimodal training. \textbf{(3)} We release \datasetname~to advance multimodal research for Africa's diverse languages and cultures.
% \begin{itemize}
%    \item We introduce \datasetname, the first large-scale multilingual visual cultural question answering dataset with questions\textit{ both in text and speech} for \textit{15 African languages} and their translations in English, spoken in 12 countries, collected and evaluated by native speakers.
%    % \item We provide detailed benchmarking of different MMLMs in different setups.
%    \item We demonstrate that \textbf{text-based multilingual ability does not transfer to speech understanding}, emphasizing the importance of increasing African language representation in multimodal model training.
%    % \item  We evaluate both open-source and proprietary %state-of-the-art 
%    % MLLMs on \datasetname~ comparing their performance across text and speech modalities, \textbf{multiple-choice} and \textbf{open-ended VQA}, and native versus English language settings.
%    \item  We open-source \datasetname~to promote research on multimodal systems that can more effectively understand and represent Africa's diverse languages and cultures.
%    % We open-source the collected dataset to support the development of future systems that better understand cultures and languages across Africa. 
% \end{itemize}
\section{Related work}
With the growing popularity of LLMs and MLLMs, cultural and multilingual multimodal evaluation has received greater attention.  For instance, \citet{liu2021visually} tested models on verifying statements about pairs of culturally related images, but this was framed merely as a binary classification task. CVQA \cite{romero2024cvqa} and CulturalVQA \cite{nayak2024culturalvqa} explicitly target cultural knowledge through human-written questions, yet they are limited to a small set of languages per continent. ALM-bench \cite{vayani2025all} expands the language coverage, but relies heavily on LLM-generated questions and web-sourced images. Additionally, a common limitation across these benchmarks is that they query models only through text, overlooking the importance of speech, an important modality for communities where language is mostly spoken.

% \paragraph{Evaluations for African Languages and Regions}
Previous work on African languages has focused mainly on text-based benchmarks. While some multilingual resources, such as GlobalMMLU \cite{singh2024global_mmlu}, include a small subset of African languages, other region-specific benchmarks provide broader coverage. Examples of these include MasakhaNEWS \cite{adelani-etal-2023-masakhanews} that evaluates text classification, AfriQA \cite{ogundepo2023afriqa} for cross-lingual question answering, or larger suites such as AfroBench \cite{ojo2025afrobench} and IrokoBench \cite{adelani-etal-2025-irokobench} that cover multiple \texttt{text} tasks. Across these efforts, a consistent finding is that models perform poorly on African languages, with open-weight models showing a significant gap to proprietary systems.

Despite this progress, all of these evaluations remain text-only, while visual knowledge and multimodal reasoning for African contexts are still largely untested. Existing multi-region multimodal datasets \cite{romero2024cvqa, vayani2025alm} include only a few African languages and do not adequately capture the region's diversity.
\datasetname~ addresses these gaps with three key contributions: (\textbf{1}) broader coverage with 15 languages across 12 countries, (\textbf{2}) inclusion of speech modality with parallel native and African-accented English audio, and (\textbf{3}) diagnostic control experiments that separate linguistic competence from cultural knowledge limitations. 
% These enable analysis of \textit{why} models fail, beyond measuring \textit{whether} they fail.
% To address this, we introduce \datasetname, the first human-created benchmark designed to evaluate multimodal cultural knowledge in Africa, querying through both text and speech. Our dataset covers 15 African languages, includes native-collected questions grounded in local imagery, and enables evaluation of vision-language and audio-visual LLMs. Unlike prior work, \datasetname~ directly addresses the needs of African languages that are primarily spoken and underrepresented in existing multimodal benchmarks.

\section{Dataset}
% In this section, we describe the languages covered, our data collection strategy, the annotation pipeline, and quality assurance procedures. We followed the data collection approach of \citet{romero2024cvqa} to collect Afri-MQA, adding \textbf{Audio QA} and expanding the number of questions per language and the number of African languages.

To create \datasetname, we selected 15 widely spoken languages in sub-Saharan Africa (by number of speakers, according to Ethnologue\footnote{\url{https://www.ethnologue.com/insights/ethnologue200/}})
%. While prior multimodal evaluation frameworks cover only a few African languages, \datasetname~ spans multiple language families and 
across 12 countries, representing an aggregate speaker population of approximately 392.6 million (see Table~\ref{tab:languages}). 
% removed table below -- too much space
\begin{table*}[h!]
\begin{center}
\small
\scalebox{0.9}{
%\resizebox{\linewidth}{!}{
\begin{tabular}{@{}lll|rrrl@{}}
\toprule
\textbf{Lang-Country} & \textbf{Family / Branch}       & \textbf{Reg}  & \textbf{\#spk} & \textbf{\#QA} & \textbf{\#eng(h)} & \textbf{\#nat(h)} \\ \midrule
Akan/Twi-Ghana & Niger-Congo / Volta-Niger    & West          & 9M                   & 537           & 2.41                                                                & 2.43                                                                       \\
Amharic-Ethiopia & Afro-Asiatic / Ethio-Semitic & East          & 57M                  & 500           & 1.56                                                                & 1.51                                                                       \\
Chichewa-Malawi & Niger-Congo / Bantu           & S \& E & 12M                  & 501           & 1.41                                                                & 1.50                                                                        \\
Hausa-Nigeria & Afro-Asiatic / Chadic        & West         & 77M                  & 496           & 2.80                                                                 & 3.04                                                                       \\
Igbo-Nigeria & Niger-Congo / Volta-Niger    & West          & 31M                  & 501           & 1.61                                                                & 1.59                                                                       \\
Kikuyu-Kenya& Niger-Congo / Bantu           & East          & 8.1M                 & 495           & 1.66                                                                & 1.72                                                                       \\
Kinyarwanda-Rwanda & Niger-Congo / Bantu          & East         & 18M                  & 501           & 2.73                                                                & 2.67                                                                       \\
Luganda-Uganda      & Niger-Congo / Bantu          & East    & 10M   & 500 & 2.30 & 2.42 \\
Oromo-Ethiopia & Afro-Asiatic / Cushitic      & East          & 37M                  & 512           & 2.20                                                                 & 2.30                                                                        \\
Setswana-Botswana& Niger-Congo / Bantu    & South         & 14M                  & 502           & 1.89                                                                & 2.39                                                                       \\
Somali-Somalia & Afro-Asiatic / Cushitic      & East          & 22M                  & 501           & 1.96                                                                & 1.99                                                                       \\
Tigrinya-Eritrea & Afro-Asiatic / Ethio-Semitic & East          & 9M                   & 537           & 2.13                                                                & 2.31                                                                       \\

Yoruba-Nigeria& Niger-Congo / Volta-Niger    & West          & 46M                  & 498           & 1.98                                                                & 2.06                                                                       \\

Sesotho-Lesotho    & Niger-Congo / Bantu           & South         & 13.5M                & 533           & --                                                                   & 1.90                                                                        \\
Zulu-S.Africa       & Niger-Congo / Bantu          & South         & 28M                  & 528           & --                                                                  & 1.37                                                                       \\ \bottomrule
\end{tabular}
}
% \vspace{-2mm}
\caption{\textbf{Overview of languages in \datasetname}. \#spk = estimated L1 \& L2 speakers, \#eng (h) = hours of accented English audio, \#nat (h) = hours of native language audio. -- indicates English audio not collected for that language.}
\label{tab:languages}
\end{center}
\vspace{-4mm}
\end{table*}
% \paragraph{Annotator selection}
We hired native language speakers as annotators through Upwork.\footnote{https://www.upwork.com/} The selection criteria were based on (i) fluency in English, (ii) prior annotation or data collection experience, (iii) high project completion rate on Upwork, and (iv) residence in a country where the target language is spoken. 
%We hired annotators who fulfill these criteria and who are based in one of the African countries. 
After the selection process, we divided the annotation into two phases designed to ensure quality.

% Below, we (i) describe our two-phase data collection pipeline, outlining the steps followed in the annotation process and our quality assurance procedures, and (ii) describe the structure of each data point (image, text, and audio).% together with statistics on language and regional coverage. 

% \subsection{Collection and Annotation Pipeline}

\subsection{Dataset collection}

\paragraph{Guideline and Platform Preparation}
We followed the annotation guidelines of \citet{romero2024cvqa}, including image categories, question templates, and distractors, and extended them to include audio recording instructions (for both native language and English) and adding a two-step review and verification process. Full annotation guidelines are provided in Appendix~\ref{annotation_guideline}.
\paragraph{Training and Screening Phase}
The first phase consisted of a small-scale pilot designed to train and screen annotators. We provided detailed guidelines and training to ensure annotators understood the task criteria and quality standards. After training, each annotator submitted 50 QA samples. We reviewed these submissions to verify alignment with the guidelines. Based on this review, we provided detailed feedback and, when necessary, scheduled follow-up meetings to clarify issues. Only annotators whose submissions met our quality criteria and successfully incorporated feedback were selected for the next phase. Approved samples from this screening phase were included in the final dataset.
% to ensure the quality of the data collection. Therefore, we began by providing them with guidelines and training to ensure they understood the criteria and quality standards of the task. After training, they were tasked with submitting 50 QA samples and we reviewed these submissions to verify that the task understanding and quality were in line with the provided guidelines. After the review, we provided detailed feedback and, when necessary, scheduled follow-up meetings to review their submissions and clarify any issues. Only annotators whose submissions met our quality criteria and incorporated the feedback successfully were selected for the next phase. 
%If the submitted data do not meet our quality criteria or indicate that the annotator has not fully understood the task, we replace them with new annotators.

\paragraph{Main Annotation Phase}  
% This was the main data collection stage, during which 
In this phase, the remaining 450 items per language were collected by annotators selected in the first phase.
To ensure quality across the large volume of submissions, we involved \textit{language coordinators}, experienced native speakers with strong linguistic and cultural knowledge for each language. Coordinators reviewed all submissions for linguistic accuracy, cultural appropriateness, adherence to guidelines, and audio quality.
% They were tasked with reviewing all submissions for linguistic accuracy, cultural appropriateness, adherence to guidelines, and audio quality. 
Review guidelines are provided in Appendix~\ref{review_guideline}. When issues arose, coordinators discussed them directly with annotators to resolve them. Unresolved disagreements were then raised with the project team for a  final decision. \textit{Language coordinators} are co-authors of this paper and played a major role in the quality assurance process. After their approval, the project team conducted a final review to ensure overall data quality.

% \begin{figure*}[ht]
%     \centering
%     \includegraphics[width=\linewidth]{latex/images/data_collection_process2.pdf}
%     \caption{Our Data collection pipeline has two phases. Phase 1 acts as a screening round for annotators, submitting only 50 images.}
%     \label{fig:placeholder}
% \end{figure*}

\subsection{Dataset Composition}

Each data point in \datasetname~ includes an image and a set of carefully constructed multiple-choice questions in both text and speech modalities, both in English and the native language. Below, we describe each of these components.

\paragraph{Image / Category Selection}  

% Images provide the visual context from which models are expected to answer questions. 
To build the image set for \datasetname, we encouraged annotators to contribute their own images whenever possible. When self-sourcing was not feasible, we permitted the use of open-license images from websites we provided (see Appendix \ref{annotation_guideline} for list of websites). All collected images were categorized into the 10 classes defined by \citet{romero2024cvqa}. Figure \ref{fig:image_cat_dist} shows the distributions of images per category (See Appendix \ref{image_cate} for category distributions across languages).
% Figure \ref{fig:cate_per_lang} presents these categories together with their distribution across languages.

\begin{figure}[h!]
    \centering
    \includegraphics[width=\linewidth]{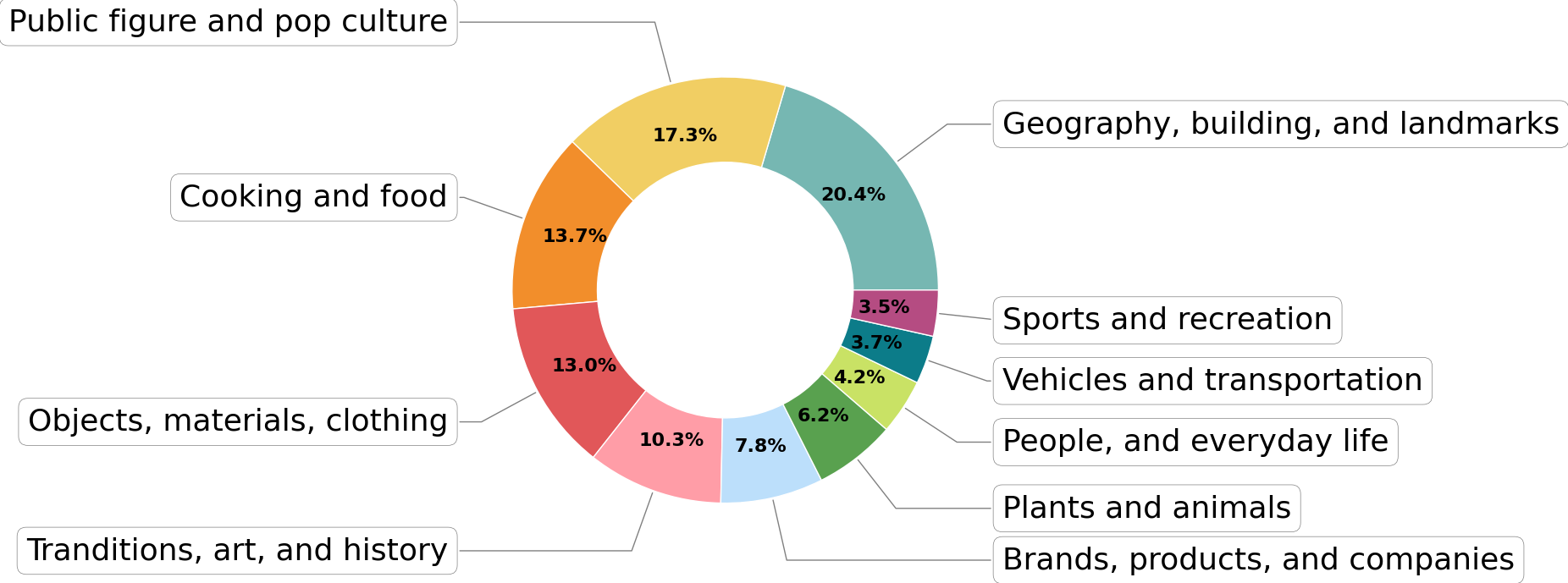}
    \caption{Image categories in our dataset and their distributions. 
    % Geography, buildings, and landmarks (580) and public figures and pop culture (491) are the most represented.}
    }\label{fig:image_cat_dist}
    \vspace{-4mm}
    
\end{figure}
\paragraph{Question \& Answer Generation}  
For each image, annotators wrote up to 3 multiple-choice QA triplets (question + 1 correct answer + 3 distractors) in both their native language and English. To ensure the benchmark remains challenging, we instructed annotators to design complex questions that require reasoning to answer correctly (See annotation guidelines in Appendix \ref{annotation_guideline}).  
% To create a strong benchmark dataset, we requested annotators to design complex questions that require reasoning to be correctly answered 

\paragraph{Audio Recording} 
To investigate spoken language understanding capabilities, we instructed annotators to record audio for each question and its corresponding answers, reading them clearly in both the native language and English. Therefore, \datasetname~ includes audio recordings for both questions and answers in native language and African-accented English.

% To enable speech-based evaluation, \datasetname~ includes audio recordings of all questions and answers in both native languages and African-accented English, recorded by annotators.
\section{Experimental setup}
Our experimental design is organized to address each research question as follows:  
% Below, we describe the models, query modalities, and query languages used in our evaluation.  
%add short paragraph here 
\paragraph{Models:} To assess how well current MLLMs understand African cultural contexts (RQ1), we selected models based on two criteria: a) support for both image and audio input, enabling multimodal evaluation, and (b) availability of different model sizes within each family to assess scaling effects. 
% Due to computational constraints, we focus on smaller model variants, which also reflects the practical reality of deploying models in resource-limited settings.
% We evaluate a variety of multimodal open-weight models for a comprehensive evaluation of \datasetname. Specifically, 
From open-weight MLLMs, we selected  Qwen 2.5-Omni (3B \&7B) \cite{Qwen2.5-Omni} and Gemma-3n-(2B \& 4B)-it \cite{gemmateam2025gemma3technicalreport}. For text-only baselines, we include Gemma3 (12B \& 27B)-it \cite{gemmateam2025gemma3technicalreport}. For comparison with closed-source models, we include
Gemini-2.5 Pro \cite{comanici2025gemini}, which supports audio, text, and vision inputs.

\paragraph{Query Modality:} To explore how input modality affects performance (RQ2), we evaluate models using both text and audio modalities. For text evaluation, models receive the written version of the question. For audio evaluation, we use native speaker recordings of the same questions and options, allowing us to assess how well current MLLMs handle spoken language inputs both in African languages and accented English. This comparison shows whether model performance generalizes from text to speech. Since we evaluate VQA, all settings include the image related to the question.

\paragraph{Query Languages:} To explore how query languages affect models' performance and whether gaps reflect linguistic or cultural limitations (RQ3), each question is presented in \textit{native language} and \textit{English}. This setup allows us to compare model behavior between English and native languages.

\paragraph{Task Format:} To understand how task format affects model performance (RQ4), we evaluate models on both Multiple-Choice VQA (MC-VQA) and Open-ended VQA. MC-VQA provides answer options, while Open-ended VQA requires answer generation. Comparing these formats shows whether strong MC-VQA performance reflects actual cultural understanding or simply selecting from provided options. For each task format, we use the same prompt templates across models and languages.    
% \paragraph{Visual Question Answer (VQA) format} We consider two evaluation formats for each modality. \textbf{For Text-based VQA}: \textit{Multiple-Choice VQA (MC-VQA)} and \textit{Open-ended VQA}.  
% In MC-VQA, models are asked to select the correct answer from four options (A–D), allowing us to directly measure accuracy.  
% In Open QA, models are asked to generate the correct answer without providing options, which tests their ability to retrieve or reason about cultural knowledge without clues.  
% \textbf{Audio-based VQA}:\textit{Multiple-Choice VQA (MC-VQA)} and \textit{Audio Answer Generation}.  

% \subsection{Prompt}
\paragraph{Prompt:} We evaluated all settings and models using a \textit{Location-aware} prompt (adding location/country as a context). We chose this setting as it performed better than image-only prompts  (without providing context). We provide results for Image-only prompts and language-wise results in Appendix~\ref{sec:add_experiment}, and the prompts used are in Appendix~\ref{prompt}.

\section{Evaluation}
% To answer our research questions, we design
% We use two evaluation setups: \textbf{(i)} cultural VQA evaluation using \datasetname ~to assess multimodal cultural understanding, and \textbf{(ii)} controlled experiment to disentangle language understanding from cultural knowledge gaps.
This section describes our two evaluation setups and the metrics used in this study. 
% Section~\ref{cultural_vqa} presents cultural VQA evaluation on \datasetname, Section~\ref{control} presents controlled experiments on established benchmarks and Section \ref{metrics} presents evaluation metrics we used in this study.
% Visual Question Answer (VQA) Evaluation Formats} 
\subsection{Cultural VQA Evaluation} \label{cultural_vqa}
We evaluate models on visually-grounded cultural QA using \datasetname ~in both modalities. For all tasks, models are provided with an image and must use visual information to answer the question. 
% We evaluate models across two modalities (text and audio) using two distinct formats for each, enabling a comprehensive assessment of multimodal understanding capabilities.
\paragraph{Text-based VQA:} For the text modality, we use two evaluation formats:
\textbf{(1) MC-VQA}: Models are given an image, a text question, and four answer options. They must select the correct option by reasoning over the image and question. \textbf{(2) Open-ended VQA}: Models receive an image and a text question without answer options, and are required to generate the correct answer. This tests their ability to retrieve and reason about cultural knowledge without potential hints in the answer set.
\paragraph{Audio-based VQA:} Similarly, we evaluate audio modality using two formats:
\textbf{(1) Audio MC-VQA}: Same setting as MC-VQA described above, but models are queried through African-accented English and native language speech. \textbf{(2) Audio open-ended VQA}: Given an image and the question in speech format without answer options and models required to generate the correct answer in text.

% This allows us to compare performance across modalities (text vs. audio) and response formats (selection vs. generation), showing how models handle different input types and whether they can understand African-accented English and native African languages across various task formulations.

% \subsection{Controlled Experiments} \label{control}
\subsection{Control Experiments} \label{control}
% Language and Speech Understanding}

While it is technically challenging to determine whether prediction failures on \datasetname~ arise from limitations in language understanding or gaps in cultural knowledge, we conduct control experiments on easy tasks that primarily require language understanding in either text or speech form. These evaluations provide evidence to understand the extent to which language-understanding limitations may contribute to the observed failures.

\paragraph{Text-based experiments:} To probe text understanding, we evaluate on two benchmarks: \textbf{(1) AfriXNLI} \cite{adelani-etal-2025-irokobench}: natural language inference, and \textbf{(2) AfriMMLU} \cite{adelani-etal-2025-irokobench}: general knowledge QA. By analyzing performance on these text-only tasks and comparing it with results on \datasetname, we obtain an approximate measure of the models' baseline linguistic competence on the studied languages.

\paragraph{Audio-based experiments:} To probe audio understanding, we conduct two tasks: \textbf{(1) ASR:}  transcribing spoken African language audio to text, assessing whether models can accurately capture spoken content as a prerequisite for answering questions. \textbf{(2) Language Identification (LID):} identifying which of the 15 languages is spoken, testing the model's ability to recognize spoken languages. These tasks reveal whether poor audio VQA results from speech-processing failures or cultural reasoning limitations.

% In the result section, we report results for the location-aware setup as our primary evaluation; comprehensive results for Image-grounded and language-wise results are provided in Appendix \ref{sec:add_experiment}. We also present the prompts used for experiments in Appendix \ref{prompt}. 

\subsection{Evaluation metrics} \label{metrics}
We evaluate models in a zero-shot setting using automatic metrics across all tasks, with additional human evaluation for open-ended VQA (text).
%We perform zero-shot evaluation of the models in both QA formats. location-aware prompt

\textbf{Automatic Evaluation:} We report accuracy scores for MC-VQA and classification tasks, and use GPT-4o-mini \cite{hurst2024gpt} as a judge for Open-ended VQA.
%, labeling each prediction as \textit{Match} (semantically equivalent to gold) or \textit{Wrong} (incorrect or irrelevant). We report accuracy as the fraction of Match responses. 
For Open-ended QA, we additionally compute chrF++ \cite{popovic2015chrf} scores and present them in Appendix~\ref{sec:add_experiment}. We report Word Error Rate (WER) for ASR.

\textbf{Human Evaluation:} We evaluate 50 randomly sampled questions per language on the best-performing model per family. Bilingual native speakers rated whether model outputs matched the gold answer or were valid alternatives.
% \textbf{(I) Automatic metrics}  We report the accuracy score for MC-VQA, whereas for Open-ended VQA, we use LLM as a judge (we also report chrf++ \cite{popovic2015chrf} score in the Appendix \ref{sec:add_experiment} as additional metrics for Open-ended VQA). We use GPT-4o-mini \cite{hurst2024gpt} as a judge for LLMs-as-judge evaluation, where each model prediction is compared against the reference and labeled under a two-tier schema: \textit{Match} (semantically equivalent to the gold answer), or \textit{Wrong} (factually incorrect or irrelevant). Only responses labeled as \textit{Match} are considered valid responses to compute accuracy.
    
% \textbf{(II) Human Evaluation.} Given the cultural and multilingual nature of \datasetname~, we additionally conducted human evaluations on 50 randomly sampled questions per language. These were conducted on the best-performing model in each family for Open-ended VQA with Location Aware prompt. 

% Each language was assigned one rater, a bilingual native speaker, under the assumption that the task requires cultural familiarity. Raters were presented with the image, the question, nine answer candidates, and the ground-truth answer (explicitly marked). They were instructed to select all answers that were either exact matches to the gold or valid alternative correct answers.  We report human accuracy as the fraction of gold answers correctly identified, and compute correlations with automated scores on the same subset to assess alignment.

\section{Results}
In this Section, we present results organized by evaluation type, beginning with cultural VQA performance across MC-VQA and Open-ended VQA tasks, followed by control experiments.
 % We present our results by modality, beginning with the text-based tasks and later extending to audio-based evaluations. For both, we report results for Multiple-Choice (MC-VQA) and Open-ended VQA in English and native languages. 
 %Per-language scores can be seen in \color{red} Appendix \color{black}
\subsection{Cultural VQA Results}
We show evaluations on MC-VQA and Open-ended VQA tasks, comparing performance across text and audio modalities in both English and native languages.
\subsubsection{Text-based QA}
\begin{figure*}[htbp!]
    \centering
    \subfigure[MC-VQA (Text)]{%
        \includegraphics[width=0.49\textwidth]{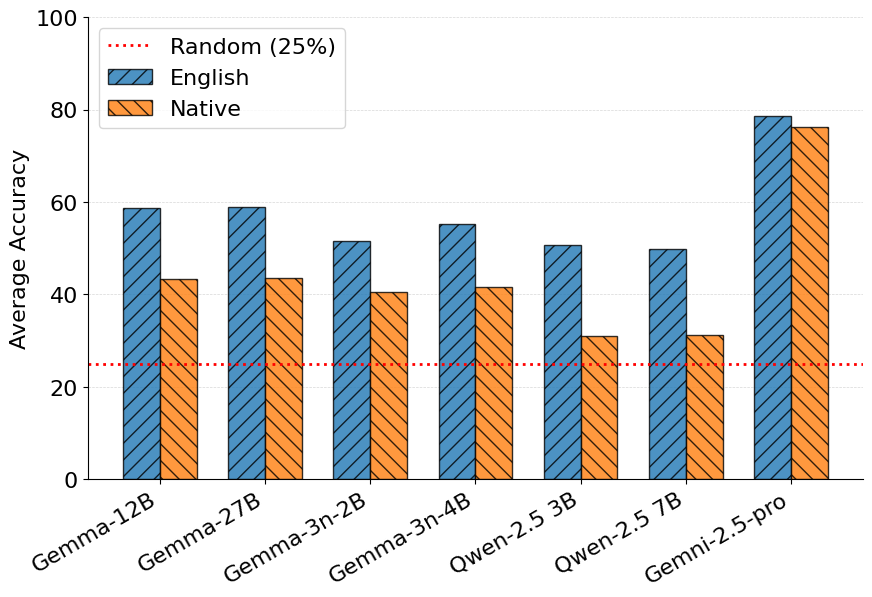}
        \label{fig:pre-mcqa}
        }
    \hfill
    \subfigure[Open-ended VQA (Text)]{%
        \includegraphics[width=0.49\textwidth]{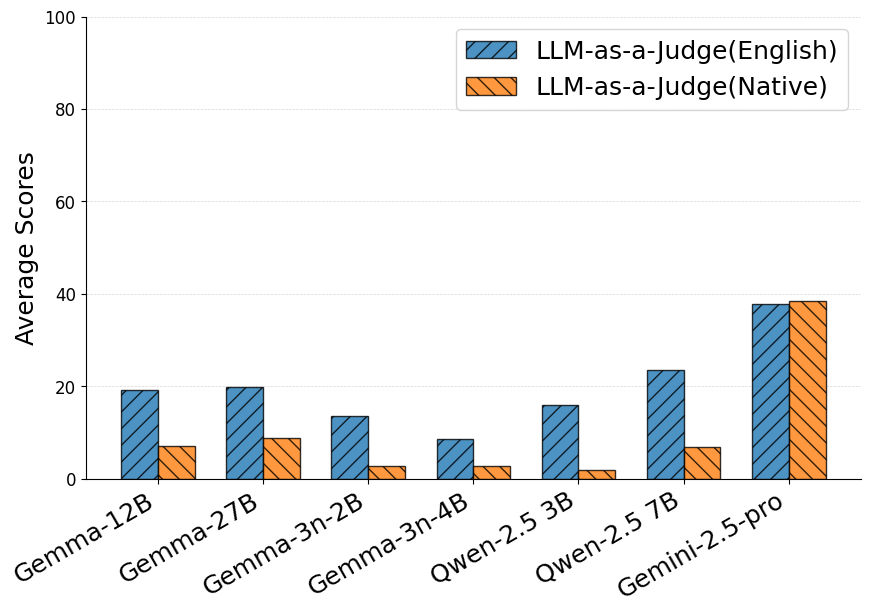}
        \label{fig:pre-open}
    }
    \caption{Performance comparison of models on text-based question answering tasks: (a) Text MC-VQA (Multiple Choice) and (b) Text Open-ended QA in English and Native languages.}

    \label{fig:avrg-text-performance}
    \vspace{-4mm}
\end{figure*}

% \paragraph{MC-VQA}
Figure \ref{fig:avrg-text-performance} depicts MC-VQA average accuracy (\ref{fig:pre-mcqa}) and LLM-as-a-judge scores (\ref{fig:pre-open}) for open-ended QA in both English and native African languages using location-aware prompts.

\begin{figure*}[ht!]
    \centering

    \subfigure[MC-VQA (Audio)]{%
         \includegraphics[width=0.49\linewidth]{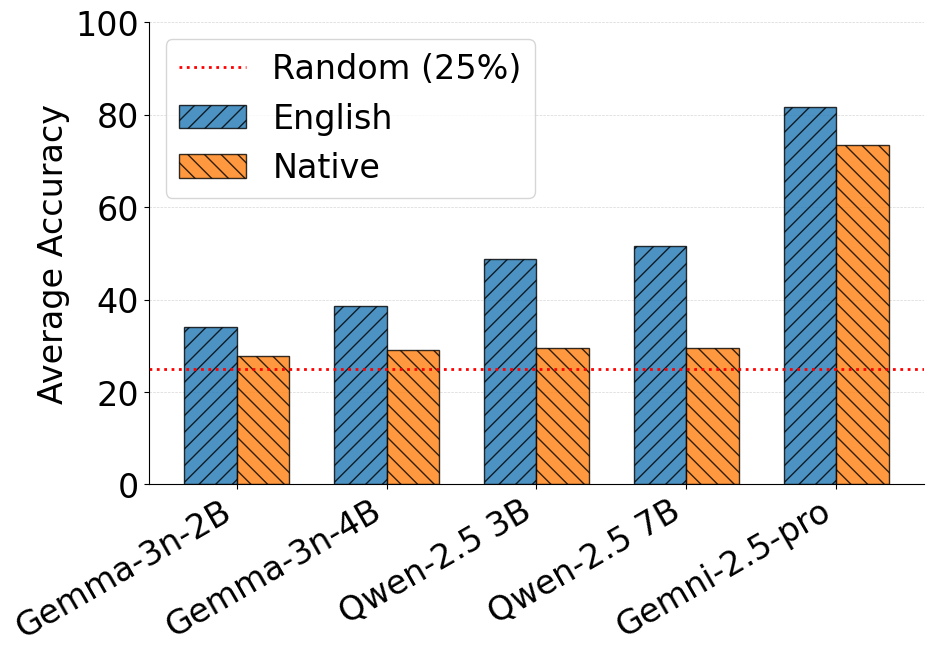}
         \label{fig:audio_mcqa}
       
    }
      \hfill
    \subfigure[Open-ended VQA (Audio)]{%
          \includegraphics[width=0.49\linewidth]{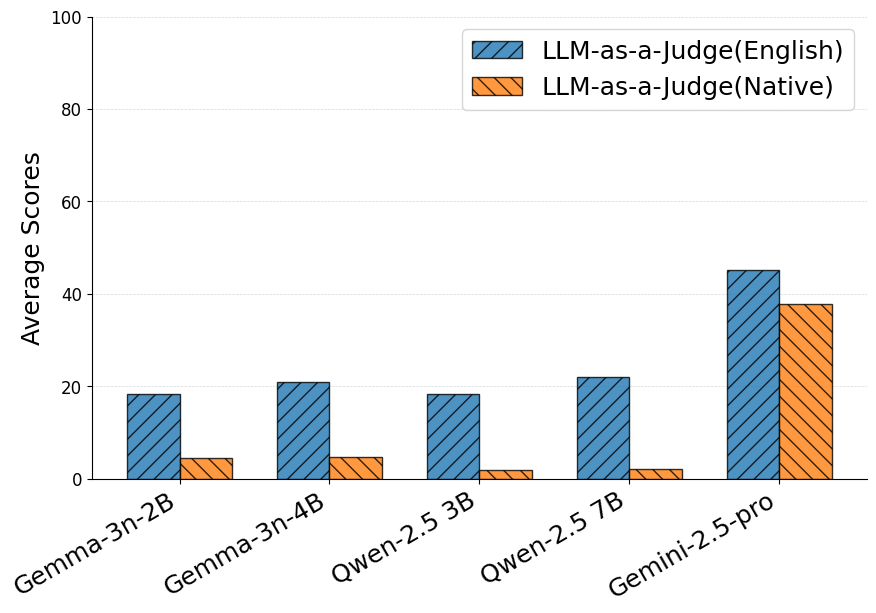}
        \label{fig:audio_oepn_ended}
        }
    \caption{Performance comparison of models on audio-based question answering tasks: (a) Audio MC-VQA (Multiple Choice) and (b) Audio Open-ended VQA in English and Native languages.}
    \label{fig:avrg-audio-performance}
    \vspace{-4mm}
\end{figure*}
% \begin{figure}[ht]
%     \centering
%     \includegraphics[width=\linewidth]{latex/images/MC-VQA_average.pdf}
%     \caption{Average MC-VQA accuracy for seven multimodal LLMs evaluated in English and in native African languages}
%     \label{fig:mcq_average}
% \end{figure}

% As expected, we find that o
Open-weight models consistently perform better when the question is in English compared to native languages across both tasks. We also observe a significant performance gap between MC-VQA and Open-ended QA, where all models, including Gemini-2.5 Pro, show major drops when tasked with answering in an open-ended setting, even in English. This suggests that \textit{generating culturally grounded responses is more challenging than selecting from predefined options}, particularly for native-language queries, where performance degrades significantly compared to English queries.
% For MC-VQA, we observe a performance gap of 11-15\%  for Gemma model family and 18-19\% for Qwen's. Gemini-2.5 Pro outperformed allnat open-weight models by a significant margin in both native and English QA, while showing a much smaller performance gap (2.5\%) between them. 

\begin{figure}[h!]
    \centering
    \includegraphics[width=\linewidth]{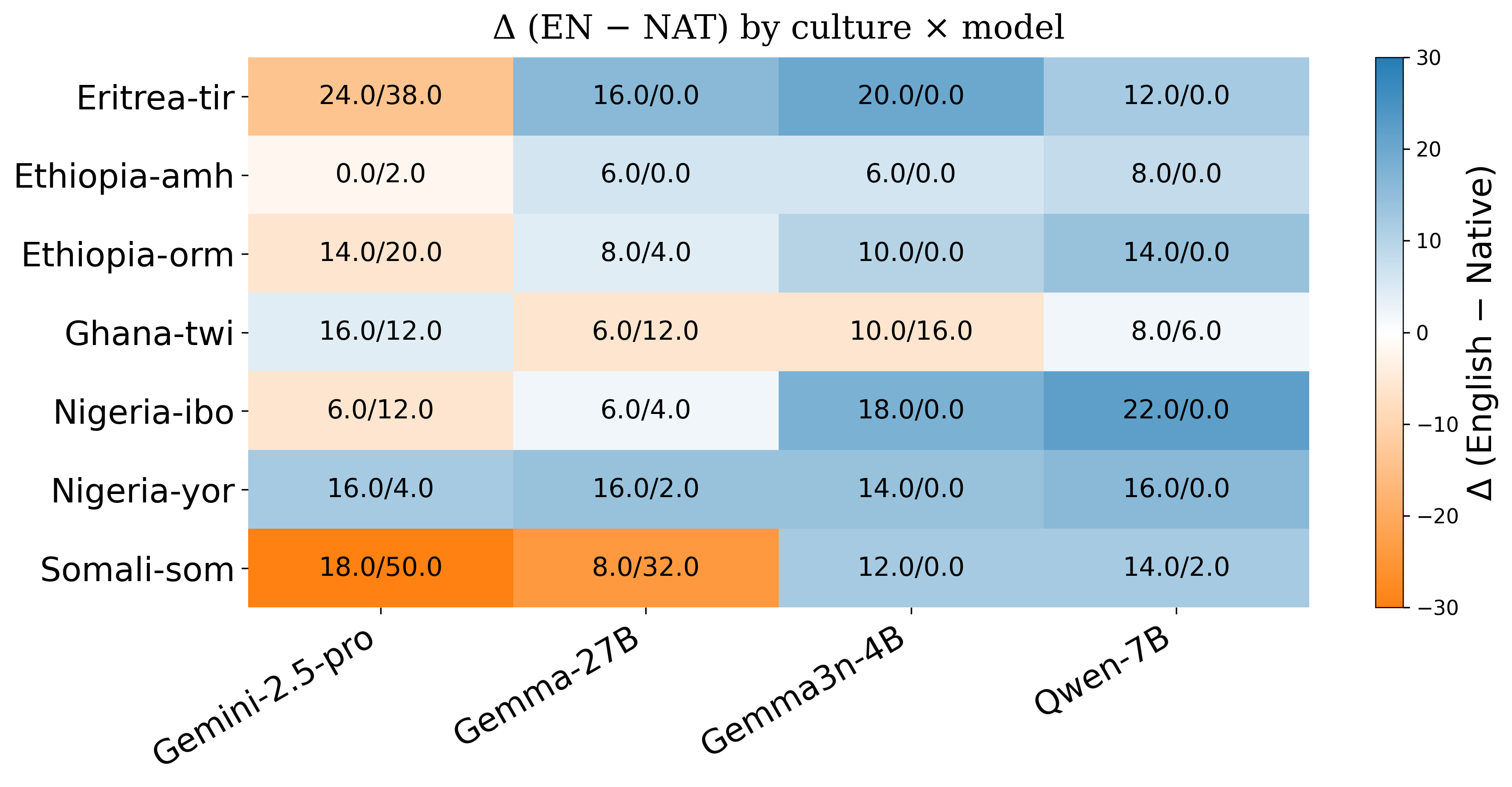}
    \caption{\textbf{Human Evaluation for Text Open-ended VQA.} Accuracy across best-performing models for English and Native. We observed that, while most models perform best in the English setting, Gemini-2.5 Pro seems to perform better in the Native language.}
    \label{fig:humaneval_heatmap}
    \vspace{-4mm}
\end{figure}

We also observe that increasing model size among open-weight models does not necessarily translate to improved performance in low-resource languages. Larger variants show limited or no improvement over smaller counterparts in native language QA, indicating that model scaling alone is insufficient to address low-resource challenges. Notably, some smaller models achieve near-zero accuracy for native languages for Open-ended QA. In contrast, Gemini-2.5-Pro outperforms all open-weight models across both tasks while maintaining comparable performance between English and native languages, \textit{highlighting the current gap between proprietary and open-weight models}. 
Human evaluations on Open-Ended VQA (in Figure \ref{fig:humaneval_heatmap}), conducted on a random subset of 50 samples across eight languages, are consistent with the trends observed in automatic evaluations, with 
%shows that all open-weight models perform better in English, with some smaller models even achieving near-zero accuracy in native languages. Consistent with the automatic evaluations, 
Gemini-2.5 Pro performs the best when queried in native languages compared to English.

% Unlike MCQA, we observe a positive correlation between model size and performance, except for the Qwen models. Scaling  Gemma model from 12B to 27B showed slight gains (19.1 to 19.8 on English  and 7.1 to 8.9 on native language in LLM-as-a-Judge score). Similarly, we observed larger improvement in the Gemma-3n family, where the 4B model visibly outperformed the 2B --smaller-- variant. Similar to MC-VQA, closed-source model Gemini-2.5-Pro achieved scores of 38.4 for English tasks, with Gemma-3n-4B coming second with scores of 28.4 for English.

% We observed that small models (<4B) performed particularly poo in the native languages, consistently scoring below 8 across all metrics. This suggests that there may be a potential minimum parameter size for effective multilingual language generation. 

\subsubsection{Audio-based QA}
Figure \ref{fig:avrg-audio-performance} shows MC-VQA accuracy (\ref{fig:audio_mcqa}) and LLM-as-a-judge scores for Open-ended QA (\ref{fig:audio_oepn_ended}) on audio inputs in both English and native African languages. 
% shows the performance of five multimodal LLMs on audio MC-VQA (\ref{fig:audio_mcqa}) and audio answer generation (\ref{fig:audio_oepn_ended}) tasks, evaluated in both accented African English and native African languages using location-aware prompts. 
Similar to text-based evaluation, open-weight models achieve higher performance when queried in English compared to native languages across both tasks. For open-weight models, \textit{audio modality is significantly more difficult than text modality}, with notable performance degradation across both tasks. Gemini-2.5-Pro, however, demonstrates robust multimodal capabilities, with consistent performance across modalities.  The performance drop from MC-VQA to Open-ended QA is also noticeable in the audio modality, with nearly zero accuracy in spoken native languages. These results align with our LID and ASR analyses (Section \ref{auio_probing}). Open-weight models demonstrate poor LID capabilities, especially the Qwen variants, which exhibit near-random accuracy and significantly compromised native ASR performance compared to English. These failures also affect performance models in downstream tasks, such as Open-ended VQA. Similar to text evaluations, we find that scaling up model size shows little improvement in native language understanding. 

% \subsection{Control Results}
\subsection{Results for Control Experiments}
We present the results on control experiments on established benchmarks alongside our cultural QA task, aiming to probe the linguistic competence of the tested models and provide pointers on why models fail on \datasetname.

\subsubsection{Text-based Experiments}
In Table \ref{tab:text_probing}, we report the performance of the models on the control-experiment benchmarks in addition to \datasetname. All models showed performance degradation from English to native languages across all benchmarks. Gemini-2.5-Pro maintains the smallest gaps, particularly on Afri-MCQA, where the drop is minimal. Open-weight models show a big drop, with Qwen variants showing the most severe gaps on AfriXNLI and AfriMMLU.  

When comparing AfriMMLU with Afri-MCQA, open-weight models show considerably higher AfriMMLU scores in English than Afri-MCQA scores, suggesting that \textit{while models possess general factual knowledge, they lack Africa-specific cultural understanding}. Gemini-2.5-Pro shows a smaller gap between these benchmarks, indicating it has acquired more African cultural knowledge during training. AfriXNLI exposes the most severe cross-lingual gaps, particularly for Qwen models, suggesting that linguistic reasoning tasks are more sensitive to language resource availability than factual retrieval tasks.

We additionally compute Spearman rank correlations between \datasetname~ and our text control datasets (see Appendix \ref{correlation_stat}). We observe strong and statistically significant correlations between AfriXNLI and AfriMMLU performance for several models. In contrast, correlations between \datasetname~ and AfriXNLI or AfriMMLU are generally weaker and not statistically significant. Hence, while limitations in language understanding may play a large role in explaining \textit{why} models fail in \datasetname, other factors related to African cultural and visual knowledge may be important as well.

\begin{table}[h!]
\centering
\small
\resizebox{\linewidth}{!}{
\begin{tabular}{lccccccccc}
\toprule
& \multicolumn{3}{c}{\textbf{AfriXNLI}} & \multicolumn{3}{c}{\textbf{AfriMMLU}} & \multicolumn{3}{c}{\textbf{Afri-MCQA}} \\
\cmidrule(lr){2-4} \cmidrule(lr){5-7} \cmidrule(lr){8-10}
\textbf{Model} &Eng& Nat & $\Delta$&Eng & Nat & $\Delta$ &Eng& Nat & $\Delta$ \\
\midrule
Gemini-2.5-Pro &89.66& 76.3 & \textcolor{red}{-13.36} &94& 83.46 & \textcolor{red}{-10.54} &78.68& 76.27 & \textcolor{red}{-2.4} \\
Gemma3n-4B &78.33& 51.24 & \textcolor{red}{-27.09} &62.2& 37.56 & \textcolor{red}{-24.64} &55.23& 41.49 & \textcolor{red}{-13.7} \\
Gemma3n-2B &80.66& 51.47 & \textcolor{red}{-29.19} &53.6& 35.74 & \textcolor{red}{-17.86 }&51.54& 40.32 & \textcolor{red}{-11.2} \\
Qwen-7B &81.11& 34.33 & \textcolor{red}{-46.78} &73.8& 36.10 & \textcolor{red}{-37.70} &49.75& 31.30 & \textcolor{red}{-18.46} \\
Qwen-3B &65.66& 36.7 & \textcolor{red}{-28.96} &65.8& 34.72 & \textcolor{red}{-31.08}& 50.68& 31.00 & \textcolor{red}{-19.65} \\
\bottomrule
\end{tabular}
}
\caption{Text-based performance (\%). Eng = English accuracy,  Nat = Native accuracy, 
$\Delta$ = English $-$ Native gap (negative = drop).}
\label{tab:text_probing}
 \vspace{-4mm}
\end{table}

\subsubsection{Audio-based Experiments} \label{auio_probing}
\begin{figure}[h!]
    \centering
    \includegraphics[width=\linewidth]{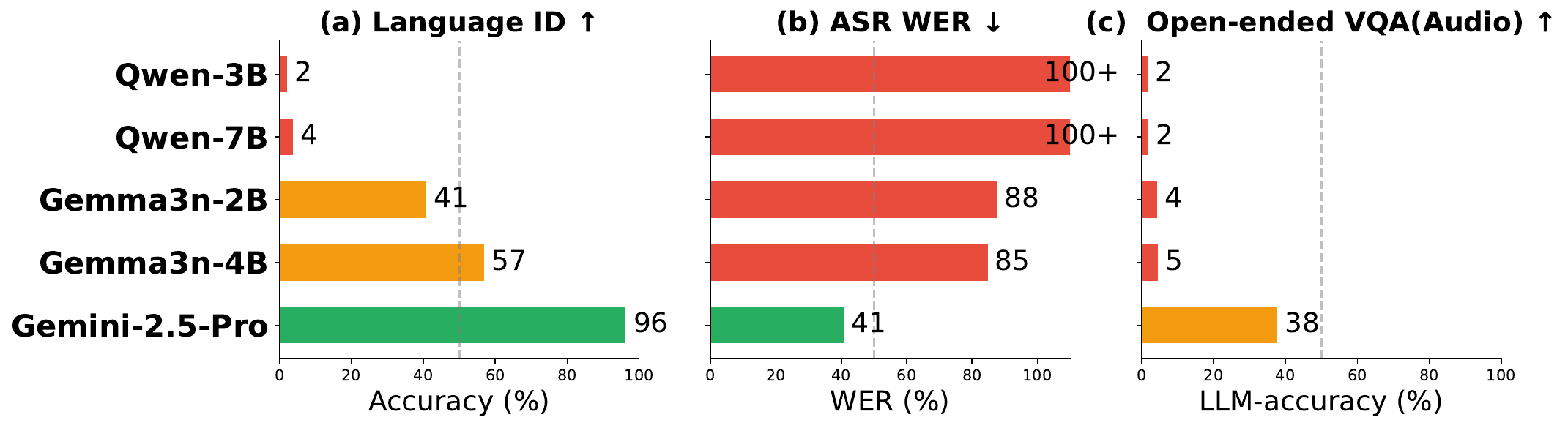}
    \caption{Audio probing results on native African languages: (a) LID  ($\uparrow$ higher is better), (b) ASR ($\downarrow$ lower is better), and (c) Open-ended VQA ($\uparrow$ higher is better). ($+$) means WER is more than 100\%}
    \label{fig:audio_prob}
     \vspace{-4mm}
\end{figure}

Figure~\ref{fig:audio_prob} shows three evaluations on native African language audio. \textbf{LID:} Gemini-2.5-Pro achieves near-perfect LID accuracy, while Gemma variants show moderate performance. Qwen models perform at near random levels, indicating minimal exposure to African language audio during pretraining. \textbf{ASR Performance:} WER shows a similar pattern. Gemini-2.5-Pro maintains reasonable native ASR, whereas Gemma models exhibit substantial degradation. Qwen models produce high error rates, indicating hallucinations rather than meaningful transcription.  This aligns with observed results for open-ended VQA (Audio). Gemini-2.5-Pro achieves moderate accuracy, whereas open-weight models score near zero. In light of these results, \textit{it is likely that models fail on open-ended VQA (Audio) because they cannot properly identify or transcribe the tested spoken languages}.

%\paragraph{Failure Analysis:} 

Hence, poor Open-ended VQA (Audio)  results come from errors at each step: \textbf{(1)} Open models often fail at identifying African languages, reflecting fundamental gaps in audio representation, \textbf{(2)} they show high ASR errors, suggesting that most spoken content is not perceived adequately for understanding and answering a question \textbf{(3)} these errors can carry over to open-ended VQA when models receive wrong transcriptions, as they cannot answer correctly even if they know the cultural content. These findings demonstrate that \textit{foundational speech processing capabilities are prerequisites for meaningful evaluation of cultural reasoning in African languages}.

\section{Discussion }
We now organize our findings to address each research question, summarizing key patterns observed across models, modalities, and languages.
\paragraph{\colorbox{rq1color} {\color{white}RQ1\color{black}}: How well do MLLMs understand African cultural contexts in visually-grounded QA?}

MLLMs show limited understanding of African cultural contexts. As shown in Figures \ref{fig:avrg-text-performance} and \ref{fig:avrg-audio-performance}, even the best performing model (Gemini-2.5-Pro) achieves only 78\% on MC-VQA and 38\% on Open-VQA (text-based, English). Smaller models (Gemma3, Qwen2.5) perform substantially worse, ranging from 50--59\% on MC-VQA and 9--24\% on Open-VQA, indicating significant room for improvement.

\paragraph{\colorbox{rq2color} {\color{white}RQ2\color{black}}: How does input modality (text vs. speech) affect performance?}

Performance degrades when switching from text to speech input, particularly for smaller models. As shown in Figures \ref{fig:avrg-text-performance} and \ref{fig:avrg-audio-performance}, on MC-VQA, Qwen models drop by approximately 1--2\% while Gemma models show mixed results. The gap is more visible in Open-VQA, where audio-based native language queries yield near-zero accuracy (2--5\%) for smaller models. Control experiments for audio show that this comes from poor language identification (2--4\% for Qwen) and high ASR word error rates (85--100\%+ for non-Gemini models).

\paragraph{\colorbox{rq3color} {\color{white}RQ3\color{black}}: How does query language (native vs. English) affect performance, and do differences reflect language understanding or cultural knowledge gaps?}

English queries consistently outperform native language queries across all models and settings. As shown in Figure \ref{fig:avrg-text-performance}, the gap ranges from 2\% (Gemini-2.5-Pro on MC-VQA) to 19\% (Qwen on MC-VQA). Control experiments on AfriXNLI and AfriMMLU (Table \ref{tab:text_probing}) show that language understanding gaps ($\Delta$ = 13--47\%) are substantially larger than cultural knowledge gaps ($\Delta$ = 2--19\% on Afri-MCQA), suggesting language understanding is the dominant limitation. However, models also struggle with cultural QA in English, indicating that both linguistic and cultural limitations contribute to poor performance.

\paragraph{\colorbox{rq4color} {\color{white}RQ4\color{black}}: How does task format (Multiple-Choice vs. Open QA) affect accuracy?}

As shown in Figures \ref{fig:avrg-text-performance} and \ref{fig:avrg-audio-performance}, models perform significantly better on MC-VQA than Open-VQA. Gemini-2.5-Pro achieves 78\% on MC-VQA vs. 38\% on Open-VQA (a 40\% gap). Smaller models show even larger relative drops, with some scoring less than 10\% on Open-VQA. This performance gap suggests that MC-VQA benefits from simplified answer selection, while Open-VQA exposes true limitations. Models struggle to generate culturally grounded responses even in English, with performance degrading further for native languages.

\section{Conclusion}
We introduced \datasetname, the first large-scale multilingual and multimodal benchmark for African cultural visual QA, covering 15 languages across 12 countries with over 7.5k Q\&A pairs in text and speech modalities.
% All data were human-annotated by native speakers through a multi-step quality pipeline, including annotator training, review rounds by language coordinators, and final verification for linguistic and cultural accuracy. Through these steps, we ensured that \datasetname~ provides a trustworthy and r   epresentative resource for evaluating multimodal reasoning in underrepresented languages from sub-Saharan Africa.
Our evaluation shows: 1) MLLMs struggle significantly with African cultural knowledge, 2) speech processing presents a critical bottleneck, and 3) gaps persist across languages and task formats. 
% model scaling alone does not resolve these gaps in native language understanding and .

These findings motivate several research directions: (1) \textbf{speech-first approaches:} Many African languages are primarily oral, yet current open-weight models lack basic LID and ASR capabilities for these languages; (2) \textbf{culturally-grounded pretraining:} The gap between AfriMMLU and Afri-MCQA performance suggests language data alone is insufficient; models need explicit exposure to African cultural content; and (3) \textbf{cross-lingual cultural transfer:}  Models may ``know'' cultural facts in English but cannot access them through native language queries, motivating research into cross-lingual knowledge retrieval.
%; and (4) \textbf{multimodal cultural grounding:} The significant MC-VQA to Open-VQA drop (30--40\%) indicates models struggle to generate culturally grounded responses from visual input, motivating vision-language models trained on African visual contexts.

We release \datasetname~ to provide both a benchmark and a foundation for building more inclusive, culturally aware multimodal systems that better represent African languages and cultures.

\section{Limitations}
We believe Afri-MCQA represents an important step toward more inclusive evaluation by foregrounding African languages and cultural contexts that have long been overlooked in existing benchmarks. Although the dataset spans 15 languages across 12 countries, Africa is home to thousands of languages and cultural groups, many of which remain unrepresented. Furthermore, while our question categories aim to reflect culturally grounded knowledge, culture itself is fluid, subjective, and deeply contextual. Our formulation inevitably abstracts away from finer-grained variations such as regional, generational, or community-specific differences that shape cultural understanding.
As with most human-curated datasets, potential biases in data collection remain. Annotators' backgrounds and interpretations of ‘cultural relevance' may influence the formulation of questions or the selection of images. Additionally, due to computational and financial constraints, we evaluate only a limited set of open- and closed-source models, so the reported performance gaps may not fully capture the broader landscape. Finally, Afri-MCQA is a human-curated dataset created without the involvement of LLMs and with minimal reliance on web-sourced images. Because this process is inherently time-consuming, the dataset is of moderate size and intended as an evaluation benchmark rather than a pretraining or fine-tuning resource, for which it would likely cause overfitting.

\section{Ethical Considerations}

Our work involves the collection of culturally grounded question–answer pairs in 15 African languages, annotated, spoken and reviewed by native speakers. All annotators participated voluntarily and were compensated fairly for their work in accordance with local wage standards on the Upwork platform. Before beginning annotation, contributors were informed about the goals of the project, the intended use of the dataset for research and evaluation purposes, and their right to withdraw from participation at any stage.

We took several steps to ensure cultural sensitivity and respect throughout the data creation process. Question formulation guidelines were designed to avoid harmful stereotypes, offensive content, or culturally inappropriate framing. All annotations were reviewed by language coordinators who are themselves native speakers to check for accuracy, contextual appropriateness, and respectful representation. Despite these efforts, we acknowledge that culture is deeply complex and subjective, and that our dataset may still reflect certain biases or oversimplifications.

\bibliography{custom}

\appendix
\onecolumn
\FloatBarrier
\section{Image Categories} \label{image_cate}

\textbf{A) Language-wise category distribution}
\begin{figure}[h!]
    \centering
    \includegraphics[width=\linewidth]{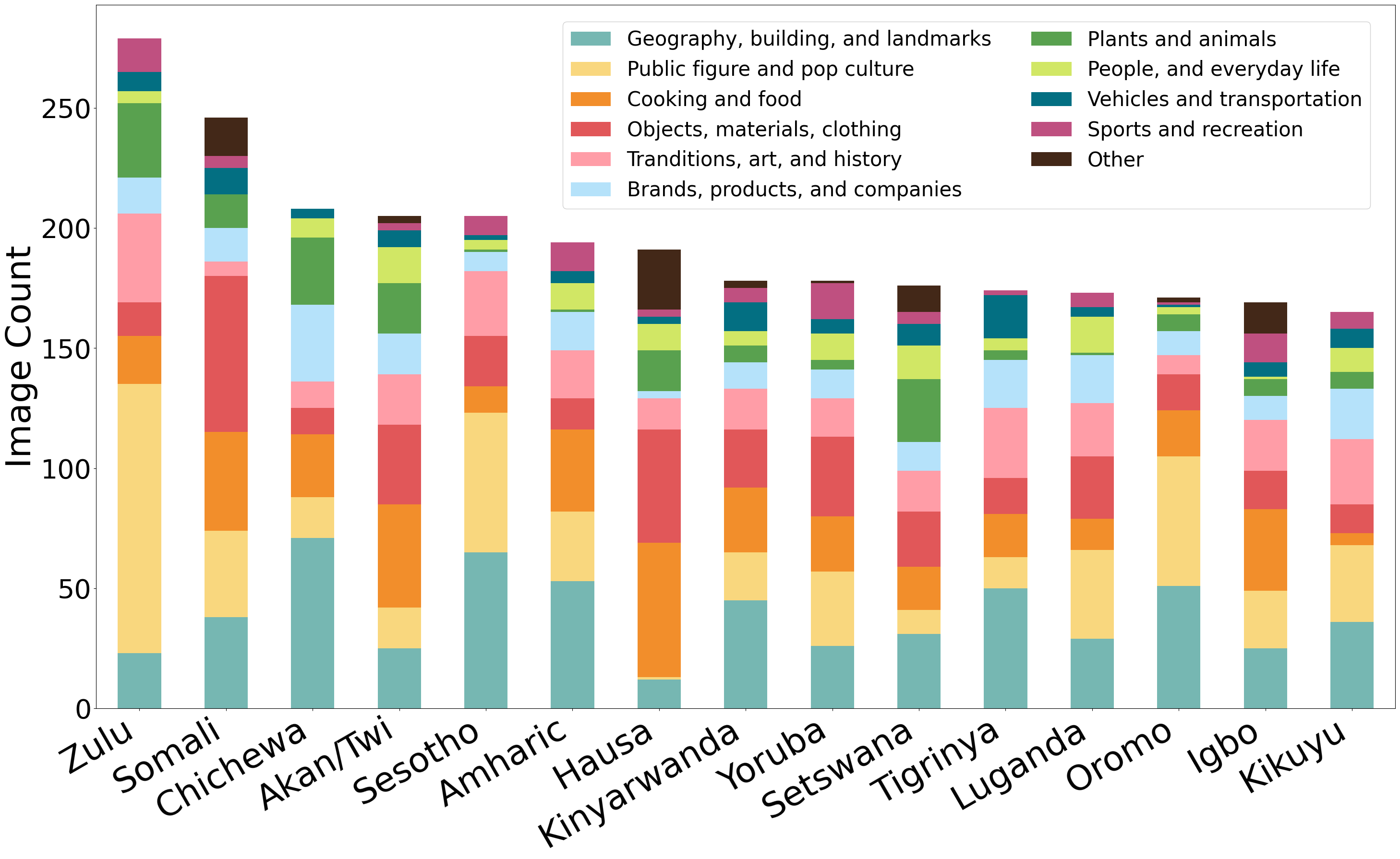}
    \caption{Language-wise distribution of the categories.}
    \label{fig:cate_per_lang}
\end{figure}

% \FloatBarrier

\textbf{B) Image category distribution}
\begin{figure*}[h!]
    \centering
    \includegraphics[width=\linewidth]{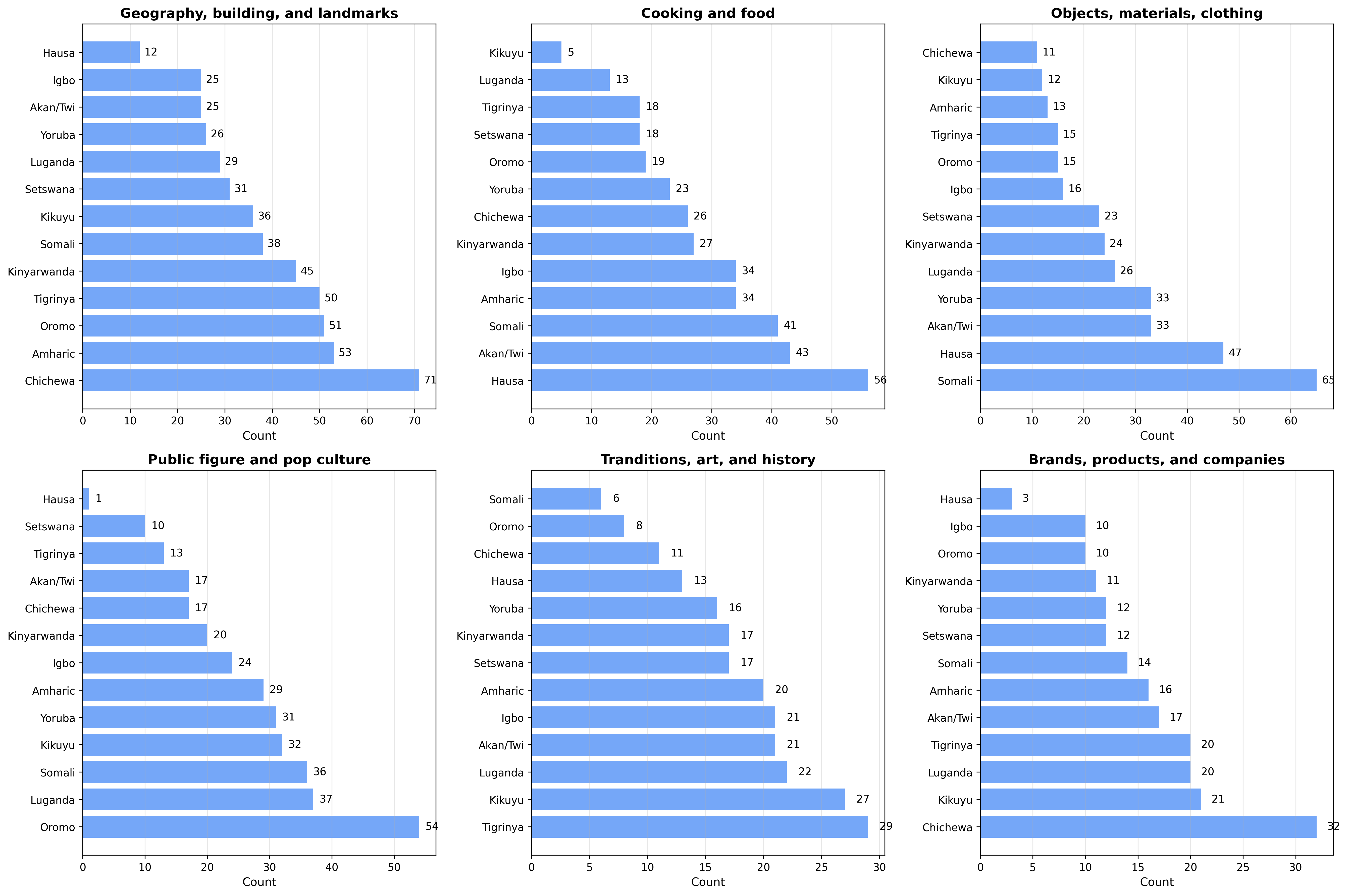}
    \caption{Top 6 Image category distribution}
    \label{fig:placeholder}
\end{figure*}
\FloatBarrier
\section{Prompt} \label{prompt}
For each QA format, we use the same prompt templates to ensure consistency across models and languages. We evaluated models under two distinct prompt conditions to assess the impact of visual and contextual grounding:
\begin{itemize}
    % \item \textbf{Text-only:} \texttt{Question: \{q\} Options: \{opts\}}
   \item\textbf{Image-grounded:} \texttt{[Image] Question: \{q\} Options: \{opts\}}
   \item \textbf{Image + Location:} \texttt{[Image] Location: \{country\}. Question: \{q\} Options: \{opts\}}
\end{itemize}
% htbp
Figures \ref{fig:text_prompts} \& \ref{fig:audio_prompts} show the location-aware prompts we used in both setups for audio and text modalities.
% \textbf{A) Location-aware prompts (text)}
\begin{figure}[htbp!]
\centering
\begin{tcolorbox}[
    enhanced,
    colback=blue!5!white,
    colframe=blue!75!black,
    title=\textbf{Text Prompts},
    fonttitle=\bfseries,
    width=\columnwidth,
    arc=2mm,
    top=3mm,
    bottom=3mm
]
\textbf{Multiple-Choice QA:}\\[0.3em]
\colorbox{gray!30}{\texttt{\tiny\bfseries SYSTEM}}\texttt{\scriptsize \ You are a helpful AI assistant.}\\[0.2em]
\colorbox{blue!30}{\texttt{\tiny\bfseries USER}}\texttt{\scriptsize \ You are given an image. Analyze the image and answer the following multiple-choice question. Only one option is correct. Return only the correct option name i.e. A, B, C or D. The question is relevant to \{country\}.}\\
\texttt{\scriptsize [Image] Question: \{question\}}\\
\texttt{\scriptsize Options: A. \{opt\_a\}, B. \{opt\_b\}, C. \{opt\_c\}, D. \{opt\_d\}}\\[0.2em]

\vspace{-0.3em}
\textbf{Open-Ended:}\\[0.3em]
\colorbox{gray!30}{\texttt{\tiny\bfseries SYSTEM}}\texttt{\scriptsize \ You are a helpful AI assistant.}\\[0.2em]
\colorbox{blue!30}{\texttt{\tiny\bfseries USER}}\texttt{\scriptsize \  You are given an image. Analyze the image and answer the question with a short factual answer. The answer should be a word or a short phrase. Only give the answer, no follow-up or extra information.The question is relevant to \{country\}}\\[0.2em]
\texttt{\scriptsize  [Image] Question: \{question\}.}\\

\end{tcolorbox}
\caption{Location-aware text prompts.}
\label{fig:text_prompts}
\end{figure}
% \textbf{B) Location-aware prompts (Audio)}  
\begin{figure}[h!]
\centering
\begin{tcolorbox}[
    enhanced,
    colback=green!5!white,
    colframe=green!75!black,
    title=\textbf{Audio Prompts},
    fonttitle=\bfseries,
    width=\columnwidth,
    arc=2mm,
    top=3mm,
    bottom=3mm
]
\textbf{Multiple-Choice QA:}\\[0.3em]
\colorbox{gray!30}{\texttt{\tiny\bfseries SYSTEM}}\texttt{\scriptsize \ You are a helpful AI assistant.}\\[0.2em]
\colorbox{green!30}{\texttt{\tiny\bfseries USER}}\texttt{\scriptsize \ You are given an image and audio question. Analyse the image and answer the audio Multiple Choice Question. Only one option is correct. Return only the correct option name i.e. A, B, C or D. The question is relevant to \{country\}.}\\
\texttt{\scriptsize [Image] [Audio] Question: \{question\_audio\}}\\
\texttt{\scriptsize Audio Options: A-D \{options\_audio\}}\\[0.2em]

\vspace{-0.3em}
\textbf{Open-Ended:}\\[0.3em]
\colorbox{gray!30}{\texttt{\tiny\bfseries SYSTEM}}\texttt{\scriptsize \ You are a helpful AI assistant.}\\[0.2em]
\colorbox{green!30}{\texttt{\tiny\bfseries USER}}\texttt{\scriptsize \ You are given an image and audio question. Analyze the image and answer the audio question with a short factual answer. The answer should be a word or a short phrase. Only give the answer, no follow-up or extra information. The question is relevant to \{country\}.}\\
\texttt{\scriptsize [Image] Question: \{question\_audio\}}
\end{tcolorbox}
\caption{Location-aware audio prompts.}
\label{fig:audio_prompts}
\end{figure}
\FloatBarrier
\section{Additional Experiments results } \label{sec:add_experiment}
\subsection{Text-based QA} - here we report additional results for different prompts and per country-language results. 
\begin{table*}[htbp!]
\centering
 \resizebox{0.95\linewidth}{!}{%
\begin{tabular}{lllllll|l}\hline
\textbf{Country-Lang} &
\textbf{G-12B} &\textbf{G-27B} & \textbf{G-3n-2B} & \textbf{G-3n-4B} & \textbf{Q-2.5 3B} &  \textbf{Q-2.5 7B} &\textbf{Gemni-2.5-pro}\\\hline
Ethiopia-amh &50.85 &50.34 & 41.23 &  40.54 &42.61   &41.23  &76.11 \\
Nigeria-hau &63.17 &62.47 &55.14 &58.21  &54.79   &57.2&81.16   \\
Nigeria-ibo &  50.19 &  49.8 &  43.72 &  40.39& 41.37  &42.15   &79.8\\
Uganda-lug &  61.65 &  62.23 &  56.92 & 60.48 &49.32   &53.75  &84.15 \\
Ethiopia-orm & 55.03 &  54.48 &  44.76 & 51.5   &45.93   &45.73  & 76.35\\
Rwanda-kin & 67.97&  68.16& 56.92 &  60.48  &55.05   &55.61  & 82.27\\
Kenya-kik  &  58.83 &  58.38 & 47.67&  54.34   &46.63   &47.67&  87.87 \\
Somali-som & 56.36 &  56.23 &   50& 54.33  &  50.27&  45.52&68.02 \\
Eritrea-tir &  57.54 &  57.89 &  45.05 & 51.22  & 50.17 & 50.52 &76.66 \\
Ghana-twi &  62.99 &  63.15 & 56.93  &   62.27 &  54.7 &  56.14 &81.33\\
Nigeria-yor  &  55.61 &  55.24 & 49.9  & 53.63  &  48.68 &  46.62 &82.95\\
Botswana-tsn &  60.41 & 61.74 &  50.94 &55.3 &  53.21 &   53.03&82.54\\
Malawi-nya &  47.91 &  47.43 &40.7   & 43.42 & 41.98 &   37.98&66.02\\
S.Africa-zul &  58.9 & 59.97 &  59.13  & 58.66 &  55.55 &51.49 &67.38\\
Lesotho-sot &  74.47 &  74.63  &  74.63 &74.47&  70.08  & 61.62&87.64\\ \hline
Average  & 58.79  &58.80& 51.54  & 55.23 &  50.68 &49.75 &78.66 \\ \hline
  
\end{tabular}}
\caption{Text Prompt - Location/Language Aware--English}
\label{tab:MC-VQA-lang-loc-english}
\end{table*}

\begin{table*}[htbp!]
\centering
 \resizebox{0.95\linewidth}{!}{%
\begin{tabular}{lllllll|l} \hline
\textbf{Country-Lang} &
\textbf{G-12B} &\textbf{G-27B} & \textbf{G-3n-2B} & \textbf{G-3n-4B}  & \textbf{Q-2.5 3B} &  \textbf{Q-2.5 7B} &\textbf{Gemni-2.5-pro}\\\hline
Ethiopia-amh &44.5 &43.64 & 42.09 & 42.78 & 30.93  &29.38  &78.17 \\
Nigeria-hau &47.12 & 49.38  & 40.31  & 42.58   &28.44  &29.12   &81.67  \\
Nigeria-ibo &  41.17 &  41.56 & 38.43 & 36.86  &   30.39 &31.56  &79.41  \\
Uganda-lug & 48.36 &  48.16 & 44.89  &  43.54 & 35.26  &35.26  &  81.31\\
Ethiopia-orm & 39.34 &  38.56 &35.65 &  36.24 &33.13   & 34.49  & 75.77\\
Rwanda-kin & 57.11& 58.05& 53.55 & 53.18 &  39.7 & 39.13 & 82.58 \\
Kenya-kik & 44.42 &  44.64 & 40.6&  41.41 &   35.25&  36.76 &85.05\\
Somali-som & 40.37&  41.19 & 43.22  & 42.68  &27.5  & 25.06 & 66.12 \\
Eritrea-tir &  47.01 &  46.84 & 41.57  &40.52 & 32.28 & 34.56 & 75.96 \\
Ghana-twi &  45.13 &  45.61 & 37.79  & 37.48  &  29.5 &29.34 &  77.51 \\
Nigeria-yor  & 44.19 & 43.82 &  38.36 & 38.36  & 31.08  & 31.27  & 81.27\\
Botswana-tsn &  39.77 & 40.15 & 36.93  & 37.87 &  30.3 & 32.38 &  83.52\\
Malawi-nya & 36.85 &  35.73 &  34.15 & 53.18  & 28.36& 28.68  & 68.58\\
S.Africa-zul & 38.23  & 39.9 &    39.18&  39.42&  27.8  & 22.46&61.88 \\
Lesotho-sot &  37.39 &  37.07  &   38.21&36.26&  25.56 & 29.91& 65.36\\ \hline
Average  & 43.39  &43.62& 40.49  &  41.49 &  31.03 &31.29 & 76.27\\ \hline
  
\end{tabular}}
\caption{Text Prompt - Location/Language Aware--native}
\label{tab:MC-VQA-lang-loc-native}
\end{table*}

\begin{table}[h!]
\centering
 \resizebox{0.95\linewidth}{!}{%
\begin{tabular}{lllllll|l}\hline
\textbf{Country-Lang} &
\textbf{G-12B} &\textbf{G-27B} & \textbf{G-3n-2B} & \textbf{G-3n-4B}& \textbf{Q-2.5 3B} &  \textbf{Q-2.5 7B} &\textbf{Gemni-2.5-pro}\\\hline
Ethiopia-amh  &50.34 &  50.51 & 42.09  &44.5 &   42.09 & 40.72 &72.23 \\
Nigeria-hau  &64.92 &65.44  & 56.36  & 57.06  & 57.59  & 58.87 & 77.31 \\
Nigeria-ibo  &48.43 &  47.05 &44.11   &47.64    & 37.64  & 40.98 &77.45 \\
Uganda-lug  & 61.65& 60.88  &  56.84 &  58.76 &   49.51 &  53.56& 80.73\\
Ethiopia-orm & 52.32&  50.58 &   46.51&  50.96 &44.18   &  44.37 & 77.32\\
Rwanda-kin  &68.53 & 67.6  &57.86   &60.48  &  54.86 &  56.92 &78.27 \\
Kenya-kik   &56.56 &55.75   & 49.09  &  54.14   &  42.02 &   47.47&86.46 \\
Somali-som  & 56.36& 57.99  &50.81   & 54.06  &  49.86  & 46.06 &64.09  \\
Eritrea-tir & 57.01&  56.49 &  41.01 & 47.89  & 48.94 &  50&  73.15\\
Ghana-twi  & 63.31& 63.15  & 59.01  &  63.95  & 55.34  &   56.45& 81.49\\
Nigeria-yor  & 55.8& 57.11  &  52.14 & 53.44 & 47.37  &  47 & 82.2\\
Botswana-tsn & 61.74& 61.17  &  53.21 & 55.11  &51.7   &  52.46& 80.8 \\
Malawi-nya & 46.15&  46.79 & 43.75  & 45.19   & 42.14  &  37.82 & 65.54\\
S.Africa-zul & 64.63  & 66.06 &   65.23&  64.99&  48.5   &53.52 &88.76 \\
Lesotho-sot &   75.28&   74.95 & 75.28  &74.63&   53.82   & 60.16& 92.5\\ \hline
Average & 58.86  &  58.76&53.28  & 55.52  &48.37& 49.75  &   78.56\\ \hline
  
\end{tabular}}
\caption{Prompt - Image only--English(text-based)}
\label{tab:MC-VQA-image-only-english}
\end{table}

\begin{table}[ht]
\centering
 \resizebox{0.95\linewidth}{!}{%
\begin{tabular}{lllllll|l} \hline
\textbf{Country-Lang} &
\textbf{G-12B} &\textbf{G-27B} & \textbf{G-3n-2B} & \textbf{G-3n-4B}& \textbf{Q-2.5 3B} &  \textbf{Q-2.5 7B} &\textbf{Gemni-2.5-pro}\\\hline
Ethiopia-amh  &43.81 &41.76   & 43.29  & 43.47   & 29.73  &   29.89 &76.8 \\
Nigeria-hau & 46.94& 46.77  & 43.45  & 44.5   &30.89   &32.16  &81.32  \\
Nigeria-ibo  &40.58 &41.76   & 38.43  & 37.64  &  30&  32.94 &79.01  \\
Uganda-lug & 50.86&50.48   &  45.08 &44.31  & 35.07  & 36.22  & 81.11 \\
Ethiopia-orm  & 39.53& 38.56  & 34.68  &  37.2 &  30.04 &  31 & 76.55\\
Rwanda-kin  &56.92 &57.67   & 55.24  &55.24  &  34.83 & 37.82 &  80.71\\
Kenya-kik   &45.85 &45.45   & 40.6  & 41.61   & 33.74  &   37.17&84.84 \\
Somali-som  &41.19 &40.78   &  43.9 & 45.12  &27.77   &24.93  &65.62  \\
Eritrea-tir  & 47.01&47.01   & 39.82  &41.05    &28.07   &  33.5 & 73.28\\
Ghana-twi  &45.45 & 45.61  & 36.52  & 31.57    &  29.18 &  29.66 & 75.91\\
Nigeria-yor  &44.19& 44  &  39.29 & 30.71  & 32.02  &  32.95&  81.23\\
Botswana-tsn  &42.04 & 40.34  &  38.06 & 31.43 & 28.4  &31.81  &  81.06\\
Malawi-nya &38.94 & 37.17  & 36.21  &35.09 & 28.68  & 28.84 &  70.19\\
S.Africa-zul & 39.66  & 39.54& 39.78   &39.3  &    29.39 & 22.46& 40\\
Lesotho-sot & 37.07  & 37.72  & 38.37  &37.2   &  28.34 &30.56 & 47.64\\ \hline
Average & 44. 00 &43.64    &41.05   &41.14&  30.27  &31.46 &73.01 \\ \hline
  
\end{tabular}}
\caption{ Image only-native (text-based)}
\label{tab:MC-VQA-image-only-native}
\end{table}

\begin{table}[ht]
\centering
 \resizebox{\linewidth}{!}{%
\begin{tabular}{l|l|l|l|l|l|l|l|l|l|l|l|l||l|l} \hline
 &
  \multicolumn{2}{l|}{\textbf{G-12B}} &
  \multicolumn{2}{l|}{\textbf{G-27B}} &
  \multicolumn{2}{l|}{\textbf{Q-2.5 3B}} &
  \multicolumn{2}{l|}{\textbf{Q-2.5 7B}} &
  \multicolumn{2}{l|}{\textbf{G-3n-2B}} &
  \multicolumn{2}{l||}{\textbf{G-3n-4B}} &
  \multicolumn{2}{l}{\textbf{Gemni}} \\
\multirow{-2}{*}{\textbf{Country-Lang}} &
  \textbf{chrf++} &
  \textbf{LLM} &
 \textbf{chrf++} &
  \textbf{LLM} &
 \textbf{chrf++} &
  \textbf{LLM} &
 \textbf{chrf++} &
  \textbf{LLM} &
  \textbf{chrf++} &
 \textbf{LLM} &
  \textbf{chrf++} &
 \textbf{LLM} &
  chrf++ &
  LLM \\\\\hline
Ethiopia-amh  & 27& 26.2 &27.55 &26.8 & 17.69  &  15 & 20.7 &  19.2 &  20.73 &16.4   &31.53   &30.4&47.88  &60.2  \\
Nigeria-hau  & 7.28& 12.63&7.47  & 13.43  & 7.73  &  7.41&  7.7 &7.82  & 10.82  & 8.45 &13.43& 11.02 & 14.63  &25.5 \\
Nigeria-ibo  & 10.41&10.54  & 11.04  &  12.13 & 10.31 &   10.14& 11.8 &  8.55 & 9.41 & 7.75  &  15.65 &15.9&  25.61&36.18  \\
Uganda-lug  &20.64 & 25.45 & 22.37  & 27.04&15.38  & 16.1  &17.51  &   18.29&17.1  & 18.89  & 27.91  &33.2&  35.7& 44.33 \\
Ethiopia-orm &18.44 & 17.19 & 19.24  & 19.34   &16.86  & 11.72  &20.08  &  17.58 &   19.2& 16.41  &  27.38 &25& 34.01&38.09   \\
Rwanda-kin  &12.51 &18.76  & 12.06 &18.36& 11.37 &  10.18 &12.51  & 12.38  &  10.82 &  16.38&16.37 & 22.92  &   18.92&36.53 \\
Kenya-kik  & 17.77&21.82  & 18.2  &  23.64  &14.42  &  11.31 &  16.54&   12.12& 16.73  &  15.15 & 30.33 &31.11 &  41.2&49.49  \\
Somali-som  &13.77 & 15.94 & 13.4  & 16.93  & 12.59 &  13.55 & 15.03 &  17.73 &   13.55&16.33 &  17.04 &   18.73& 18.62  &25.5 \\
Eritrea-tir &20.11 &28.75  &20.47   &29.29  &14.82  & 19.17  & 17.87 & 22.24  &  14.64 & 19.23  &24.22   & 30.02  &35.75&49.01 \\
Ghana-twi &20.37 & 29.05 &  21.01 & 29.24  &16.83  & 23.46  & 18.87 &   23.09& 19.41  & 24.58  &  25.71 & 32.4 &  25.71&34.64\\
Nigeria-yor   & 10.2& 17.27&  9.38&  17.87  &9.13  &13.86   &  9.92&  12.25 &10.12   & 15  &13.13   &20.6   & 20.61&43.98\\
Botswana-tsn &16.14 &20.96  & 15.16  & 19.66   &14.93  & 17.37  & 16.34 & 20.56  & 15.55  & 18.56  &  20.59 & 27.94 &22.98&28.14  \\
Malawi-nya  & 16.41& 15.17 & 15.56  &15.57  &11.08  &  17.37 & 13.39 & 12.77&  13.42 & 18.56  & 20.72  & 22.75  & 25.24&30.34\\
S.Africa-zul  & 18.48& 15.72 & 17.94  & 16.1  & 11.48 & 8.7 & 12.05 &  9.28  & 17.49  &  15.72 & 18.2  & 16.67  &34.87&39.58 \\
Lesotho-sot & 12.56& 11.26 & 14.01  &  11.82 &11.49  & 8.44 & 10.69 &  8.63  & 13.37  & 11.44  &14.16   &12.01  &24.7&25.7 \\ \hline
Average & 16.13& 19.14 &16.32   &  19.81 & 13.07 &13.58& 14.73 &  8.56& 14.61   &15.83   &20.75   &23.54   &28.43  &37.80 \\ \hline
\end{tabular}}
\caption{ Location/Language Aware-English(Open-endded VQA -text based)}
\label{tab:open_eng}
\end{table}

\begin{table}[ht]
\centering
 \resizebox{0.85\linewidth}{!}{%
\begin{tabular}{l|l|l|l|l|l|l|l|l|l|l|l|l||l|l} \hline
 &
  \multicolumn{2}{l|}{\textbf{G-12B}} &
  \multicolumn{2}{l|}{\textbf{G-27B}} &
  \multicolumn{2}{l|}{\textbf{Q-2.5 3B}} &
  \multicolumn{2}{l|}{\textbf{Q-2.5 7B}} &
  \multicolumn{2}{l|}{\textbf{G-3n-2B}} &
  \multicolumn{2}{l||}{\textbf{G-3n-4B}} &
  \multicolumn{2}{l}{\textbf{Gemni}} \\
\multirow{-2}{*}{\textbf{Country-Lang}} &
  \textbf{chrf++} &
  \textbf{LLM} &
 \textbf{chrf++} &
  \textbf{LLM} &
 \textbf{chrf++} &
  \textbf{LLM} &
 \textbf{chrf++} &
  \textbf{LLM} &
  \textbf{chrf++} &
 \textbf{LLM} &
  \textbf{chrf++} &
 \textbf{LLM} &
  chrf++ &
  LLM \\\\\hline
Ethiopia-amh  &2.47& 22.8 & 2.91  &  22.6  & 0.11 &  6.6 &  1.01&   4.2&  0 & 1  &0.71  & 20.4&42.17  &55.4 \\
Nigeria-hau  &4.23 &3.01  &4.31   &  3.21 & 2.79 & 0.6 &2.9   & 0.6 & 1.41 & 2.4  & 3.83 &2.6   &15.02   &30.26 \\
Nigeria-ibo  & 5.93&  5.96&  5.93 & 5.96  &   3.65& 0.99  & 4.97 &  0.99 &  0.44 &  0.99 & 5.42 & 2.98& 24.11 &33  \\
Uganda-lug  &9.48 &  9.96&  9.22 &   11.16&     5.69&   3.19&  4.27& 3.19  &  0.5 &1.79   &  8.45 &8.76&  30.33&40.44  \\
Ethiopia-orm & 6.56&6.45  &  6.35 &  5.86 &  5.26&  6.05 &  3.26& 1.76  &  0.68 & 0.59  & 6.27  &4.69& 37.92&39.84   \\
Rwanda-kin  &8.31 &9.38  &8.23   &  9.58  &4.8  &3.39   & 3.65 & 3.02  & 0.39  & 0.2  & 6.24 &7.19 &   25.27&34.13 \\
Kenya-kik   & 7.47& 3.03 &   7.82& 30.3  &5.81&  1.21 &  4.03& 0.61  & 1.84  &  1.21 & 9.31 &3.03 &  39.17&47.37  \\
Somali-som  &7.85 & 7.57 &   7.9&  7.37  & 3.72 &  1.39 &  4.42& 2.19  & 2.17  & 1.99  &5.64  & 4.78&22.59&30.48 \\
Eritrea-tir &0.85 & 3.98 &   0.95& 4.34&   0.04& 5.79  &  0.43&  5.79 &  0.01  & 0.18  & 0.13 & 14.1&27.23&37.43 \\
Ghana-twi &6.19 & 5.77 &   6.3& 6.89  &   4.92& 2.42  &  4.43&  2.61 & 1.88  & 0.74  & 6.89  & 5.21&31.63&33.89  \\
Nigeria-yor   &3.91 & 6.43 &   3.72& 5.62  &   2.25& 1  & 2.3  &1.2   & 1.22  &  0.6 & 4.3  &7.8& 22.33&46.18\\
Botswana-tsn &7.91 &  4.39&   7.11&  4.19  &  4.4& 2.79  & 4.15 & 2  & 2.04  &  0.6 & 7.26  & 5.99 &34.41&39.75  \\
Malawi-nya  & 8.99&  5.19&   8.44&  4.99 &  4.23&0.4   &  3.43&0.8   &1.71   &0.4   & 9.33  & 3.59  &31.5& 34.13\\
S.Africa-zul  & 13.54& 10.04 & 12.72  & 8.33  &  4.68& 1.33 &5.21  &4.88  & 14.32   & 10.42  &14.75   & 10.42  & 41.13&44.51\\
Lesotho-sot &7.6 &3  &7.15   & 2.81  &4.73  & 3 & 8.08 & 4.88 & 7.77 & 3.38  & 8  & 3 &28.67&29.83 \\ \hline
Average &6.75 & 7.13 & 6.60  &6.88   & 3.80 &  2.67&  3.73&  2.68& 2.42 &  1.76  &6.43   & 6.93 &30.22&38.43 \\ \hline
\end{tabular}}
\caption{Location/Language Aware-native(Open-ended VQA -text based)}
\label{tab:open-nat}
\end{table}

\begin{table}[ht]
\centering
 \resizebox{0.85\linewidth}{!}{%
\begin{tabular}{l|l|l|l|l|l|l|l|l|l|l|l|l||l|l} \hline
 &
  \multicolumn{2}{l|}{\textbf{G-12B}} &
  \multicolumn{2}{l|}{\textbf{G-27B}} &
  \multicolumn{2}{l|}{\textbf{Q-2.5 3B}} &
  \multicolumn{2}{l|}{\textbf{Q-2.5 7B}} &
  \multicolumn{2}{l|}{\textbf{G-3n-2B}} &
  \multicolumn{2}{l||}{\textbf{G-3n-4B}} &
  \multicolumn{2}{l}{\textbf{Gemni}} \\
\multirow{-2}{*}{\textbf{Country-Lang}} &
  \textbf{chrf++} &
  \textbf{LLM} &
 \textbf{chrf++} &
  \textbf{LLM} &
 \textbf{chrf++} &
  \textbf{LLM} &
 \textbf{chrf++} &
  \textbf{LLM} &
  \textbf{chrf++} &
 \textbf{LLM} &
  \textbf{chrf++} &
 \textbf{LLM} &
  chrf++ &
  LLM \\\\\hline
Ethiopia-amh  &19.86 & 15.8 & 20.04  &15.8   &10.38& 6.6 &   11.76 &  9.2 & 15.42  & 10  & 24.62 & 20.8 & 40.41 &43.8  \\
Nigeria-hau  &6.28& 12.63&6.53  &  11.62 & 7.98 & 6.8 & 7.98  & 7.01 & 7.36  &  8.02 &7.35   &12.02& 9.37  & 16.83   \\
Nigeria-ibo  &10.57 &8.55  &9.95   &8.15   &  9.78&  3.76 & 10.5 & 6.76  & 8.97  & 5.96  & 13.27  & 11.13&24.63  &30.82  \\
Uganda-lug  &20.59 & 24.25 &19.68   & 22.47  &13.85  &  13.92 &  15.89& 15.9  &  16.43 & 18.49  & 27.36&30.2    & 49.04 & 48.21 \\
Ethiopia-orm &18.79 & 15.04 & 19.01  &  14.45 & 12.29 & 6.84  & 14.68 & 11.33  & 14.08  & 8.01  & 23.32 &19.14  & 37.92&36.52   \\
Rwanda-kin  &11.41 &15.97  & 11.22  & 15.37   & 10.29 & 7.39  & 11.38 &10.18   &9.41   & 10.78  & 15.15 &18.16  &   23.07&32.93 \\
Kenya-kik  & 18.55&  20.81&18.2   & 20.2  & 12.03 & 7.68  & 15.97 &11.11   &15.12   & 13.94  &28.99   &28.48 &  49.78&59.19  \\
Somali-som  & 14.79&18.53  & 15.4  & 19.12   & 8.03 &  14.94  &12.45&  17.13 &12.97   & 15.34  & 18.04  &21.71&26.79   &32.07 \\
Eritrea-tir &14.71 & 16.09 & 14.64  & 15.19   & 8 & 6.87  & 9.84 & 7.59  & 10.71  & 7.78  & 18.09  & 12.84  & 31.98&40.14\\
Ghana-twi & 21.95&29.8  &  20.66 & 29.24    &8.85  & 18.44  &15.78  & 21.23  &26.57   &24.95   &25.57   & 33.33&36.76 &47.87  \\
Nigeria-yor   & 10.87&17.87  &  10.61 & 17.47    &7.02&  11.45 & 8.89 &11.24      & 9.25  &14.4   & 13.48  &21&23.2& 45.38\\
Botswana-tsn &15.75 & 22.95 & 16.14  &22.75    & 9.27 & 17.76  & 14.95 & 22.16  &13.77   &18.36   & 19.43  & 25.35 & 30.27&39.72 \\
Malawi-nya  &12.25 & 9.78 & 12.97  & 11.38  & 8.7 &6.19 & 9.88 & 6.19  &15.33   &8.18   &  15.33 & 12.77  &26.35&26.35 \\
S.Africa-zul  &16.12 &13.25  & 15.77  & 11.93  & 10.61 &7.1  & 10.7 &  8.52& 15.65  & 12.69  & 16.36  &  13.07 &40.02 &40.34\\
Lesotho-sot & 11.99&  9.19&  11.75 &  9.57 &  10.59& 6.94 & 10.55 &  9.01   & 12.71  & 9.94  &12.33   & 9.38 &23.31&23.83 \\ \hline
Average & 14.97&16.7  &14.84   &16.31   & 9.84 &  9.51& 12.08&11.64&13.58 & 12.46  & 19.1  &19.29   &  31.53 & 37.6 \\ \hline
\end{tabular}}
\caption{Image only-english(Open-endded VQA -text based)}
\label{tab:my-table}
\end{table}

\begin{table}[ht]
\centering
 \resizebox{0.95\linewidth}{!}{%
\begin{tabular}{l|l|l|l|l|l|l|l|l|l|l|l|l||l|l} \hline
 &
  \multicolumn{2}{l|}{\textbf{G-12B}} &
  \multicolumn{2}{l|}{\textbf{G-27B}} &
  \multicolumn{2}{l|}{\textbf{Q-2.5 3B}} &
  \multicolumn{2}{l|}{\textbf{Q-2.5 7B}} &
  \multicolumn{2}{l|}{\textbf{G-3n-2B}} &
  \multicolumn{2}{l||}{\textbf{G-3n-4B}} &
  \multicolumn{2}{l}{\textbf{Gemni}} \\
\multirow{-2}{*}{\textbf{Country-Lang}} &
  \textbf{chrf++} &
  \textbf{LLM} &
 \textbf{chrf++} &
  \textbf{LLM} &
 \textbf{chrf++} &
  \textbf{LLM} &
 \textbf{chrf++} &
  \textbf{LLM} &
  \textbf{chrf++} &
 \textbf{LLM} &
  \textbf{chrf++} &
 \textbf{LLM} &
  chrf++ &
  LLM \\\\\hline
Ethiopia-amh &4.65&	17.2&	3.78	&17.6&	3.5	&3.2&	0.68	&2.4	&1.51	&14.6&	1.21	&13.8&	42.59&	54.8 \\
Nigeria-hau & 3.42 &	3.01	 &3.83 &	2	 &3.12 &	3.41 &	3.12 &	1.4	 &3.77 &	3.81	 &3.44	 &3.81	 &14.67	 &29.66 \\
% &3.42 &  & 3.83 &   & 3.12 & & 3.12 & 1.4 &  3.03 & 1  & 3.02  && 14.67  &29.66 \\
Nigeria-ibo & 5.21	&4.17	&5.31	&2.98		&2.53&	2.78&	5.32	&0.99	&5.41	&2.58	&5.67&	2.19&	23.2	&29.62 \\
% &5.21 &  & 5.31 &   & &   &5.32  &   & 5.41  && 5.67  &   & 23.2 &29.62  \\
Uganda-lug &9.17&	10.96&	8.14	&9.96&	4.28	&4.38	&4.72&	4.58&	8.1	&8.96&	7.95&	7.57	&31.54&	39.04 \\
% &9.17 &  & 8.14  &  &  &   &4.72  &  & 8.1&   &  7.95 &   & 31.54 &39.04  \\
Ethiopia-orm&6.28	&7.03&	5.94	&6.45	&2.01	&2.15&	4.68&	2.93&	5.67&	3.32&	5.54&	2.73	&29.32	&38.48\\
% &6.28 &  & 5.94  &   & &  &  4.68  &   &  5.67 & & 5.54 &   & 29.32& 38.48  \\
Rwanda-kin  &7.66	&9.98&	7.59&	9.78&	2.36	&2.59	&3.98	&2.2	&6.69&	8.58&	6.19	&7.39&	25.51&	35.33 \\
% 7.66 &  & 7.59  &   & &   &  3.98 &   & 6.69 & & 6.19  &   &   25.51& 35.53\\
Kenya-kik  &7.58	&3.23	&7.57&	3.03&	4.53	&4.04&	6.11	&1.62&	8.7	&2.22&	8.9	&2.63&	39.23	&48.08\\
% 7.58 &  & 7.57  &    &  &   & 6.11 &   & 8.7  & & 8.9 &   &  39.23&48.08  \\
Somali-som  &7.56 & 8.17 & 7.3  &8.76   & 3.29 & 2.39  &6.21  &  3.78 &5.73   &  5.78 &  5.7 &  4.98 &25.02&32.07 \\
Eritrea-tir &1.14& 3.62 & 1.21  & 3.8  & 0.12 & 1.63  &0.19  & 0.72  &  0.74 &  4.16 &  0.69 &4.7   &27.03&35.62 \\
Ghana-twi &6.52 & 6.15 &6.24   &7.08    & 5.03 &2.42   & 5.74 & 4.28  &6.44   & 4.66  & 6.21  &4.66  & 29.44&32.22 \\
Nigeria-yor  &3.76 & 6.22 &  3.94 & 6.92   &2.3  &0.6   & 3.33 &  0.8 & 4.3  &  8.2 &4.22   & 8  &20.6&43.73 \\
Botswana-tsn &6.74 &  4.99& 6.48  &5.59     & 2.97 & 2.99  & 5.86 & 8.58  &6.6   & 4.19  &6.95   &4.59  &34.32&36.93  \\
Malawi-nya  &8.78& 5.99 &  8.3 & 1.2  & 3.31 & 1  &5.36  &2.79   &8.35   & 3.79  & 8.33  &3.19   & 29.83&31.94\\
S.Africa-zul  & 14.11& 9.66&  14.68 &10.61   &3.44  & 2.65 & 3.72   & 1.52  & 13.78  & 10.23  & 14.06  &  9.66 &40.69&42.8 \\
Lesotho-sot & 7.28& 3.75 & 7.76  & 4.69  & 3.13 & 2.25 & 6.88 & 10.69   & 7.55  & 3.19  & 7.04  &  3.19& 24.95&25.7\\ \hline
Average &6.66 &6.94  & 6.54  &6.7   & 2.97 & 2.57 &4.39  & 3.29   & 6.22  & 5.88  & 6.14  & 5.54 & 29.20&37.07\\ \hline
\end{tabular}}
\caption{Image only-native(Open-endded VQA -text based)}
\label{tab:my-table}
\end{table}

\begin{table}[ht]
\centering
\resizebox{\textwidth}{!}{%
\begin{tabular}{lccccc|ccccc}
\toprule
& \multicolumn{5}{c}{\textbf{English}} & \multicolumn{5}{c}{\textbf{Native}} \\
\cmidrule(lr){2-6} \cmidrule(lr){7-11}
\textbf{Country-Lang}
& \textbf{Gemma3n-2B} & \textbf{Gemma3n-4B} & \textbf{Qwen-3B} & \textbf{Qwen-7B} & \textbf{Gemini-2.5 Pro}
& \textbf{Gemma3n-2B} & \textbf{Gemma3n-4B} & \textbf{Qwen-3B} & \textbf{Qwen-7B} & \textbf{Gemini-2.5 Pro} \\
\midrule
Ghana-twi         & 37.78 & 44.35 & 58.96 & 62.83 & 87.01 & 28.83 & 32.11 & 31.48 & 31.27 & 69.12 \\
Ethiopia-amh    & 34.38 & 36.17 & 42.43 & 43.60 & 83.52 & 29.48 & 32.97 & 30.03 & 28.63 & 82.46 \\
Malawi-nya    & 27.60 & 33.17 & 42.96 & 44.30 & 80.09 & 26.52 & 27.98 & 29.62 & 32.08 & 70.80 \\
Nigeria-hau       & 36.96 & 41.46 & 62.74 & 62.53 & 83.61 & 27.95 & 32.29 & 29.93 & 27.51 & 82.32 \\
Nigeria-ibo        & 26.85 & 30.09 & 40.60 & 44.50 & 77.15 & 27.18 & 26.18 & 27.15 & 26.08 & 56.50 \\
Kenya-kik      & 27.12 & 28.72 & 44.41 & 45.98 & 83.28 & 26.86 & 25.00 & 31.90 & 33.15 & 72.34 \\
Rwanad-kin & 45.78 & 51.40 & 59.07 & 63.06 & 85.16 & 30.69 & 32.07 & 29.31 & 31.44 & 78.13 \\
Uganda-lug     & 39.64 & 46.10 & 47.65 & 50.56 & 81.73 & 33.87 & 31.81 & 34.39 & 38.38 & 72.48 \\
Ethiopia-orm       & 37.55 & 39.20 & 41.23 & 42.35 & 74.52 & 29.11 & 28.87 & 26.49 & 27.06 & 71.46 \\
Botswana-tsn    & 35.86 & 40.92 & 51.06 & 54.60 & 83.15 & 25.05 & 26.11 & 27.17 & 19.70 & 74.46 \\
Somali-som      & 26.24 & 30.73 & 49.17 & 56.05 & 82.14 & 24.76 & 26.42 & 24.88 & 24.14 & 75.41 \\
Eritrea-tir   & 36.55 & 44.30 & 46.98 & 52.30 & 75.48 & 28.27 & 29.76 & 31.26 & 34.33 & 72.35 \\
Nigeria-yor      & 31.17 & 35.49 & 45.56 & 47.58 & 84.13 & 23.64 & 25.53 & 26.31 & 26.68 & 76.54 \\
Lesotho-sot     & --    & --    & --    & --    & --    & 29.77 & 30.00 & 34.00 & 31.12 & 59.15 \\
S.Africa-zul        & --    & --    & --    & --    & --    & 27.04 & 26.63 & 28.89 & 31.06 & 74.18 \\
\midrule
\textbf{Average}
& \textbf{34.11} & \textbf{38.62} & \textbf{48.68} & \textbf{51.56} & \textbf{81.61}
& \textbf{27.86} & \textbf{29.01} & \textbf{29.52} & \textbf{29.51} & \textbf{73.41} \\
\bottomrule
\end{tabular}}
\caption{Audio-Loc-Lang Aware(MC-VQA)}
\label{tab:audio_loc_lang}
\end{table}

\begin{table}[!ht]
\centering
\resizebox{\textwidth}{!}{%
\begin{tabular}{lccccc|ccccc}
\toprule
& \multicolumn{5}{c}{\textbf{English}} & \multicolumn{5}{c}{\textbf{Native}} \\
\cmidrule(lr){2-6} \cmidrule(lr){7-11}
\textbf{Country-Lang} 
& \textbf{Gemma3n-2B} & \textbf{Gemma3n-4B} & \textbf{Qwen-3B} & \textbf{Qwen-7B} & \textbf{Gemini-2.5 Pro}
& \textbf{Gemma3n-2B} & \textbf{Gemma3n-4B} & \textbf{Qwen-3B} & \textbf{Qwen-7B} & \textbf{Gemini-2.5 Pro} \\
\midrule
Ghana-twi         & 33.67 & 46.40 & 57.23 & 62.19 & 87.75 & 27.61 & 24.74 & 32.55 & 32.51 & 65.16 \\
Ethiopia-amh     & 31.01 & 35.05 & 45.18 & 44.58 & 79.72 & 28.17 & 28.38 & 31.66 & 29.91 & 79.69 \\
Malawi-nya    & 25.18 & 32.44 & 41.81 & 48.50 & 77.72 & 28.71 & 29.44 & 37.07 & 26.86 & 68.87 \\
Nigeria-hau       & 36.49 & 42.18 & 63.30 & 65.41 & 84.28 & 27.95 & 30.60 & 25.24 & 24.80 & 81.36 \\
Nigeria-ibo        & 23.37 & 27.08 & 37.65 & 39.71 & 76.62 & 27.93 & 23.44 & 21.85 & 27.17 & 55.50 \\
Kenya-kik      & 25.26 & 30.31 & 45.65 & 45.96 & 83.73 & 25.53 & 25.00 & 27.74 & 35.04 & 71.81 \\
Rwanda-kin & 44.24 & 48.59 & 57.54 & 61.33 & 81.79 & 28.28 & 28.97 & 30.83 & 31.00 & 79.24 \\
Uganda-lug  & 36.74 & 44.09 & 45.49 & 45.94 & 78.65 & 32.49 & 31.12 & 35.64 & 35.20 & 69.27 \\
Ethiopia-orm       & 34.50 & 32.62 & 37.40 & 42.50 & 71.15 & 27.68 & 28.16 & 26.01 & 26.78 & 71.22 \\
Botswana-tsn    & 35.86 & 40.29 & 50.85 & 54.36 & 85.31 & 23.99 & 25.69 & 27.89 & 25.00 & 70.82 \\
Somali-som      & 24.11 & 29.78 & 49.64 & 55.34 & 80.42 & 26.65 & 22.17 & 25.30 & 24.64 & 72.64 \\
Eritrea-tir    & 33.11 & 36.34 & 47.29 & 47.67 & 74.51 & 26.98 & 27.19 & 32.10 & 31.97 & 70.39 \\
Nigeria-yor      & 23.50 & 35.25 & 47.07 & 48.92 & 85.99 & 24.82 & 25.77 & 28.01 & 26.68 & 77.96 \\
Lesotho-sot     & --    & --    & --    & --    & --    & --    & --    & 31.12 & 28.29 & 59.11 \\
S.Africa-zul        & --    & --    & --    & --    & --    & --    & --    & 29.04 & 32.09 & 72.07 \\
\midrule
\textbf{Average} 
& \textbf{31.31} & \textbf{36.96} & \textbf{48.16} & \textbf{50.02} & \textbf{80.59}
& \textbf{27.45} & \textbf{26.98} & \textbf{29.47} & \textbf{29.20} & \textbf{71.01} \\
\bottomrule
\end{tabular}
}
\caption{MC-VQA Audio - Image Only}
\label{tab:audio_image_only}

\end{table}

\begin{table}[!ht]
\centering
\resizebox{0.95\linewidth}{!}{%
\begin{tabular}{@{}lrrrrrrrrrr@{}}
\toprule
\multicolumn{1}{c}{\multirow{2}{*}{Language}} & \multicolumn{2}{c}{G3n-2B}                            & \multicolumn{2}{c}{G3n-4B}                            & \multicolumn{2}{c}{Q-3B}                              & \multicolumn{2}{c}{Q-7B}                              & \multicolumn{2}{c}{Gemini-2.5 Pro}                    \\
\multicolumn{1}{c}{}                          & \multicolumn{1}{c}{chrf++} & \multicolumn{1}{c}{acc.} & \multicolumn{1}{c}{chrf++} & \multicolumn{1}{c}{acc.} & \multicolumn{1}{c}{chrf++} & \multicolumn{1}{c}{acc.} & \multicolumn{1}{c}{chrf++} & \multicolumn{1}{c}{acc.} & \multicolumn{1}{c}{chrf++} & \multicolumn{1}{c}{acc.} \\ \midrule
Ghana-twi   & 19.21                     & 29.36                    & 21.61                     &31.01                   & 19.16                     & 29.98                   & 21.11                    & 33.06                   & 31.92                      & 45.59                   \\
Ethiopia-amh  & 17.86                     & 18.65                    & 27.61                     &24.94                  & 18.5                   & 19.05                    & 21.14                      & 22.56                  & 49.39                      & 64.04               \\
Malawi-nya  & 13.05                      & 18.16                   & 14.55                     & 20.82                    & 13.05                    & 21.22                   & 18.04                     & 24.81                    & 27.9                     & 35.35                   \\
Nigeria-hau   & 13.02                     & 14.89                    & 11.34                      & 18.68                    & 13.26                      & 14.42                    & 17.28                     & 19.76                    & 17.99                      & 31.21                    \\
Nigeria-ibo     & 12.31                      & 6.94                    & 12.09                      & 9.03                    & 11.26                      & 7.89                    & 14.48                   & 12.6                    & 30.9                      & 42.13                   \\
Kenya-kik   & 17.87                     & 14.47                    & 18.89                     & 18.16                    & 15.86                      & 14.47                   & 19.21                      & 18.23                    & 44.58                      & 55.26                    \\
Rwanda-kin     & 15.99                      & 17.14                    & 15.09                      & 17.39                    & 15.59                      & 16.25                    &16.49                     & 19.48                    & 25.82                      & 43.99                    \\
Uganda-lug     & 16.37                      & 18.71                    & 19.13                      & 21.60                    & 16.14                      & 19.6                    & 17.09                      & 20.33                    & 30.9                      & 52.34                    \\
Ethiopia-orm     & 18.72                      & 17.80                    &23.03                      & 22.25                     & 17.02                      & 15.49                    & 20.62                       & 20.54                     & 38.46                       & 45.43                     \\
Botswana-tsn    & 18.15                      & 24.10                    & 19.27                      & 25.16                    & 16.48                      & 23.33                    & 17.61                      & 23.56                    & 26.79                      & 36.81                   \\
Somali-som    & 14.39                      & 21.51                    & 16.39                      &23.17                    & 15.02                      & 18.29                    & 17.54                      & 23.99                    & 23.05                      & 32.39                   \\
Eritrea-tir     & 16.3                      & 21.08                   & 17.89                      & 21.94                    & 15.93                      & 21.59                  & 19.39                      & 27.61                    & 35.64                      & 53.33                    \\
Nigeria-yor     & 11.55                     & 16.31                    & 11.91                      & 18.71                   & 11.3                      & 16.63                    & 12.81                     & 18.99                   & 25.07                      & 49.64                   \\ \midrule
Average      & 15.75                     & 18.39                    & 17.58                      & 20.99                   & 15.27                      & 18.31                    & 17.91                      & 21.96                    & 32.11                    & 45.19                   \\ \bottomrule
\end{tabular}}
\caption{Audio Open-ended for English (Loc-Lang Aware)}
\label{tab:audio_open_ended_english}
\end{table}

\begin{table}[!ht]
\centering
\resizebox{0.85\linewidth}{!}{%
\begin{tabular}{@{}lrrrrrrrrrr@{}}
\toprule
\multicolumn{1}{c}{\multirow{2}{*}{Language}} &
  \multicolumn{2}{c}{G3n-2B} &
  \multicolumn{2}{c}{G3n-4B} &
  \multicolumn{2}{c}{Q-3B} &
  \multicolumn{2}{c}{Q-7B} &
  \multicolumn{2}{c}{Gemini-2.5 Pro} \\
\multicolumn{1}{c}{} &
  \multicolumn{1}{l}{chrf++} &
  \multicolumn{1}{l}{acc.} &
  \multicolumn{1}{l}{chrf++} &
  \multicolumn{1}{l}{acc.} &
  \multicolumn{1}{l}{chrf++} &
  \multicolumn{1}{l}{\textbf{acc.}} &
  \multicolumn{1}{l}{chrf++} &
  \multicolumn{1}{l}{acc.} &
  \multicolumn{1}{l}{chrf++} &
  \multicolumn{1}{l}{acc.} \\ \midrule
Ghana-twi        & 12.14 & 11.75 & 12.24 & 13.25 & 5.98 & 3.92 & 5.98  & 4.10 & 45.72 & 57.06 \\
Ethiopia-amh     & 0.58  & 24.85 & 1.53  & 25.85 & 0.00 & 2.61 & 0.00  & 2.61 & 47.43 & 79.30 \\
Malawi-nya    & 16.03 & 10.98 & 17.32 & 13.77 & 6.31 & 5.39 & 6.31  & 5.59 & 58.40 & 63.05 \\
Nigeria-hau       & 9.52  & 13.10 & 12.53 & 16.94 & 2.87 & 0.40 & 2.87  & 0.60 & 63.06 & 81.65 \\
Nigeria-ibo        & 9.42  & 9.90  & 9.01  & 10.71 & 4.22 & 2.83 & 4.22  & 2.83 & 41.81 & 52.08 \\
Kenya-kik      & 20.13 & 12.96 & 21.02 & 14.40 & 5.54 & 1.66 & 5.54  & 1.66 & 54.02 & 63.09 \\
Rwanda-kin & 12.81 & 14.20 & 11.84 & 17.40 & 2.48 & 1.20 & 2.48  & 1.20 & 67.48 & 73.68 \\
Uganda-lug     & 15.13 & 18.49 & 15.41 & 18.29 & 4.96 & 3.59 & 4.96  & 3.78 & 56.59 & 64.26 \\
Ethiopia-orm       & 6.22  & 10.60 & 5.80  & 9.40  & 2.90 & 2.93 & 3.27  & 3.72 & 51.83 & 50.55 \\
Botswana-tsn    & 17.03 & 15.23 & 14.58 & 12.22 & 4.88 & 3.04 & 4.88  & 3.04 & 71.22 & 74.39 \\
Somali-som      & 6.78  & 8.58  & 6.61  & 11.38 & 3.00 & 2.00 & 3.00  & 2.00 & 59.27 & 69.17 \\
Eritrea-tir    & 0.10  & 22.35 & 0.14  & 17.32 & 0.01 & 7.08 & 0.01  & 7.26 & 47.97 & 68.63 \\
Nigeria-yor      & 8.92  & 19.00 & 8.30  & 19.40 & 2.76 & 1.60 & 2.76  & 1.20 & 33.51 & 72.98 \\
Lesotho-sot     & 6.46  & 3.56  & 5.30  & 4.13  & 5.14 & 1.27 & 6.47  & 2.11 & 50.94 & 50.84 \\
S.Africa-zul        & 19.12 & 18.75 & 22.44 & 16.86 & 8.78 & 4.20 & 15.36 & 9.24 & 69.90 & 71.43 \\ \midrule
Average     & 10.69 & 14.29 & 10.94 & 14.75 & 3.99 & 2.91 & 4.54  & 3.40 & 54.61 & 66.14 \\ \bottomrule
\end{tabular}}
\caption{Audio Open-ended for Native Audio(Loc-Lang Aware)}
\label{tab:audio_native_open_ended}
\end{table}

\begin{table}[!ht]
\centering
\resizebox{0.95\linewidth}{!}{%
\begin{tabular}{@{}lrrrrrrrrrr@{}}
\toprule
\multicolumn{1}{c}{\multirow{2}{*}{Language}} & \multicolumn{2}{c}{G3n-2B}                            & \multicolumn{2}{c}{G3n-4B}                            & \multicolumn{2}{c}{Q-3B}                              & \multicolumn{2}{c}{Q-7B}                              & \multicolumn{2}{c}{Gemini-2.5 Pro}                    \\
\multicolumn{1}{c}{}                          & \multicolumn{1}{c}{chrf++} & \multicolumn{1}{c}{acc.} & \multicolumn{1}{c}{chrf++} & \multicolumn{1}{c}{acc.} & \multicolumn{1}{c}{chrf++} & \multicolumn{1}{c}{acc.} & \multicolumn{1}{c}{chrf++} & \multicolumn{1}{c}{acc.} & \multicolumn{1}{c}{chrf++} & \multicolumn{1}{c}{acc.} \\ \midrule
Ghana-twi   & 18.91                    & 28.54         &21.52           & 29.98                    &19.30                   & 29.36                     & 20.01                   & 29.55                    & 35.14                   & 57.49    \\
Ethiopia-amh  & 13.41                    &10.56                    & 15.65                     &13.26                  & 13.42                   & 11.24                    & 14.90                     & 15.04                 & 40.36                    & 49.44               \\
Malawi-nya  & 11.83                      & 9.44                   & 11.71                     & 9.69                   & 12.57                    &14.58                 & 13.30                     & 14.13                    & 25.41                     & 32.69                 \\
Nigeria-hau   & 11.58                     & 13.95                   & 11.02                      & 13.48                   & 12.67                      & 14.89                    & 15.60                     & 15.73                    & 23.14                      & 37.35                  \\
Nigeria-ibo     & 11.36                      & 4.86                    & 11.66                      & 6.25                    & 12.72                      & 5.79                    &12.72                  & 6.02                    & 26.90                      & 34.95                   \\
Kenya-kik   & 16.33                     & 13.68                    & 16.77                     & 16.58                    & 15.60                      & 13.68                   & 18.32                     & 17.24                 & 44.25                      & 57.11                    \\
Rwanda-kin     & 14.19                      & 13.81                    & 13.96                     & 12.02                    & 15.71                      & 14.87                   &15.41                     &14.92                    & 25.34                      & 33.5                   \\
Uganda-lug     & 14.81                      & 16.7                    &16.32                      & 19.38                   & 15.25                      & 19.15                   &15.76                      & 18.18                    & 37.54                     & 49.22                    \\
Ethiopia-orm     & 14.36                      & 9.6                    &17.3                      & 11.94                    & 13.85                      & 10.07                    &15.75                       & 14.67                    &38.42                      & 42.62                    \\
Botswana-tsn    & 15.82                     & 20.93                    & 17.02                      & 23.89                    & 14.70                      & 22.03                    &18.99                      & 24.6                    & 34.18                      & 49.89                   \\
Somali-som    & 13.62                      &19.39                    & 14.7                     &22.22                    & 14.63                      & 20.33                   & 16.56                      & 22.8                    &25.76                      & 37.35                   \\
Eritrea-tir     & 12.36                     & 8.39                   & 12.33                     & 10.11                    & 12.37                      & 8.62                  & 12.62                    & 10.87                    & 29.77                      & 39.57                    \\
Nigeria-yor     & 10.67                   &13.91                   & 10.65                      & 14.39                   & 10.81                     & 13.43                    & 11.40                    & 17.46                  & 27.21                      & 49.4                 \\ \midrule
Average      & 13.79                     & 14.13    &14.66                & 15.63                      & 14.12                   & 15.25                      & 15.49                   & 17.01                      & 31.80                   & 43.89   \\ \bottomrule
\end{tabular}}
\caption{Audio Open-ended for English (Image only Audio)}
\label{tab:audio_open_ended_english}
\end{table}

\begin{table}[ht]
\centering

\resizebox{\textwidth}{!}{%
\begin{tabular}{lcccccccccc}
\toprule
& \multicolumn{10}{c}{\textbf{Native}} \\
\cmidrule(lr){2-11}
\textbf{Language}
& \multicolumn{2}{c}{\textbf{G3n-2B}}
& \multicolumn{2}{c}{\textbf{G3n-4B}}
& \multicolumn{2}{c}{\textbf{Q-3B}}
& \multicolumn{2}{c}{\textbf{Q-7B}}
& \multicolumn{2}{c}{\textbf{Gemini-2.5 Pro}} \\
\cmidrule(lr){2-3} \cmidrule(lr){4-5} \cmidrule(lr){6-7} \cmidrule(lr){8-9} \cmidrule(lr){10-11}
& chrF++ & LLM-acc
& chrF++ & LLM-acc
& chrF++ & LLM-acc
& chrF++ & LLM-acc
& chrF++ & LLM-acc \\
\midrule
Ghana-twi         & 6.96 & 2.45 & 6.89 & 4.70 & 9.83 & 6.75 & 9.99 & 5.52 & 15.72 & 18.40 \\
Ethiopia-amh     & 0.08 & 3.28 & 0.15 & 6.11 & 6.74 & 1.31 & 5.27 & 1.09 & 28.23 & 58.52 \\
Malawi-nya    & 8.44 & 2.68 & 7.75 & 2.43 & 9.68 & 3.65 & 9.38 & 2.43 & 20.66 & 21.90 \\
Nigeria-hau       & 7.45 & 4.58 & 7.35 & 3.86 & 10.69 & 3.86 & 9.85 & 4.58 & 22.09 & 36.63 \\
Nigeria-ibo        & 6.39 & 0.25 & 5.51 & 1.00 & 7.23 & 0.50 & 6.64 & 0.75 & 11.23 & 15.25 \\
Kenya-kik      & 9.07 & 2.13 & 9.14 & 2.93 & 9.03 & 2.13 & 9.09 & 1.86 & 20.14 & 23.73 \\
Rwanda-kin & 8.11 & 5.17 & 7.55 & 5.17 & 9.84 & 5.86 & 9.47 & 3.79 & 25.57 & 41.03 \\
Uganda-lug     & 8.21 & 4.35 & 7.51 & 5.49 & 5.94 & 2.75 & 5.26 & 1.37 & 24.11 & 31.65 \\
Ethiopia-orm       & 9.04 & 6.44 & 8.46 & 8.11 & 4.67 & 0.72 & 8.15 & 0.95 & 27.87 & 40.19 \\
Botswana-tsn    & 6.97 & 2.12 & 6.89 & 3.40 & 9.69 & 4.67 & 9.89 & 3.82 & 24.55 & 31.62 \\
Somali-som      & 6.65 & 4.25 & 5.81 & 7.55 & 11.33 & 10.38 & 11.52 & 10.14 & 17.59 & 29.48 \\
Eritrea-tir    & 0.02 & 3.21 & 0.04 & 4.28 & 4.67 & 0.43 & 5.35 & 0.00 & 15.71 & 36.70 \\
Nigeria-yor      & 4.18 & 1.42 & 4.00 & 3.31 & 6.84 & 2.36 & 7.39 & 2.84 & 15.73 & 35.63 \\
Lesotho-sot     & 9.01 & 2.89 & 8.41 & 4.22 & 8.70 & 1.56 & 8.94 & 2.00 & 13.68 & 20.67 \\
S.Africa-zul        & 9.65 & 5.12 & 10.54 & 6.76 & 7.88 & 2.66 & 7.67 & 2.25 & 34.04 & 40.37 \\
\midrule
Average& 6.68 & 3.36
& 6.40& 4.62& 8.18 & 3.31& 8.26 & 2.89& 21.13 & 32.12 \\
\bottomrule
\end{tabular}}
\caption{Audio Open-ended for Native (Image only Audio)}
\label{tab:audio_open_ended_native}
\end{table}
\subsection{Language wise heatmap for MCQA-text}
\begin{figure*}[h!]
    \includegraphics[width=\linewidth]{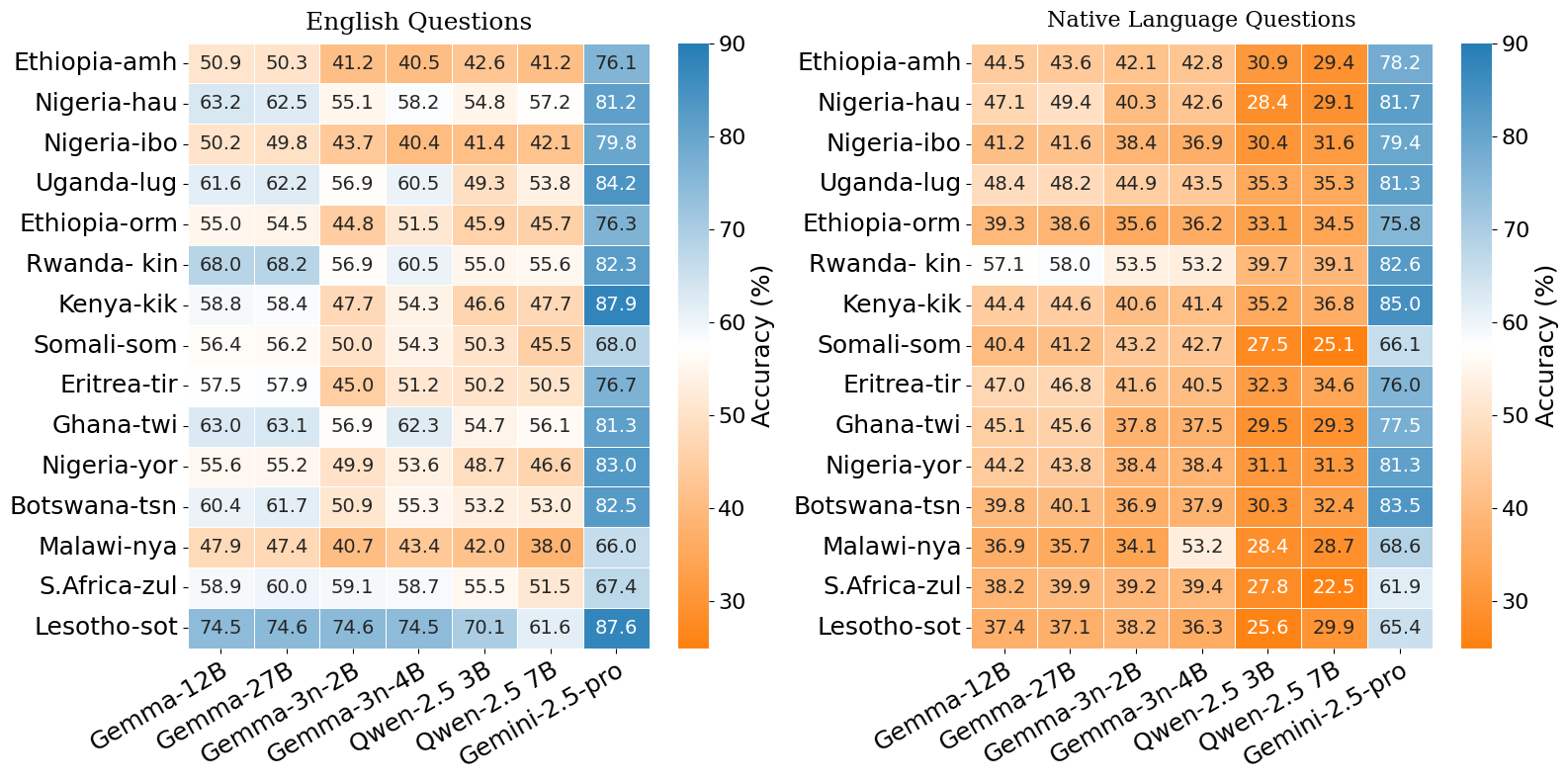}
    \caption{Language-wise accuracy of seven multimodal LLMs on the text-based VQA task, evaluated separately on English and native African language questions.}
    \label{fig:heatmap(MCQA)}
\end{figure*}

\section{Correlation statistics} \label{correlation_stat}
\begin{table}[]
\centering
\resizebox{\textwidth}{!}{%
\begin{tabular}{llllll}
% \resizebox{\textwidth}{!}{%
\toprule
 & Gemma3n-2B & Gemma3n-4B & Qwen2.5-3B & Qwen2.5-7B & Gemini-2.5 Pro \\
\midrule
XNLI - MMLU      & 0.684 (p=0.029) & 0.88 (p=0.001)  & 0.254 (p=0.479)  & 0.426 (p=0.22)  & 0.831 (p=0.003)  \\
XNLI - Afri-MCQA & 0.261 (p=0.498) & 0.636 (p=0.065) & -0.388 (p=0.302) & 0.316 (p=0.407) & -0.248 (p=0.519) \\
MMLU - Afri-MCQA & 0.373 (p=0.288) & 0.463 (p=0.178) & 0.236 (p=0.511)  & 0.479 (p=0.162) & -0.222 (p=0.537) \\
\bottomrule
\end{tabular}%
}
\caption{}
\label{tab:correlations}
\end{table}

\FloatBarrier
% % \section{Results per category}
% \section{Annotation guideline} \label{annotation_guideline}
% \includepdf[pages=-]{Updated guidelines for Annotators.pdf}
% \FloatBarrier

\includepdf[
  pages=-,
  pagecommand={
    \section{Annotation guideline}\label{annotation_guideline}
  }
]{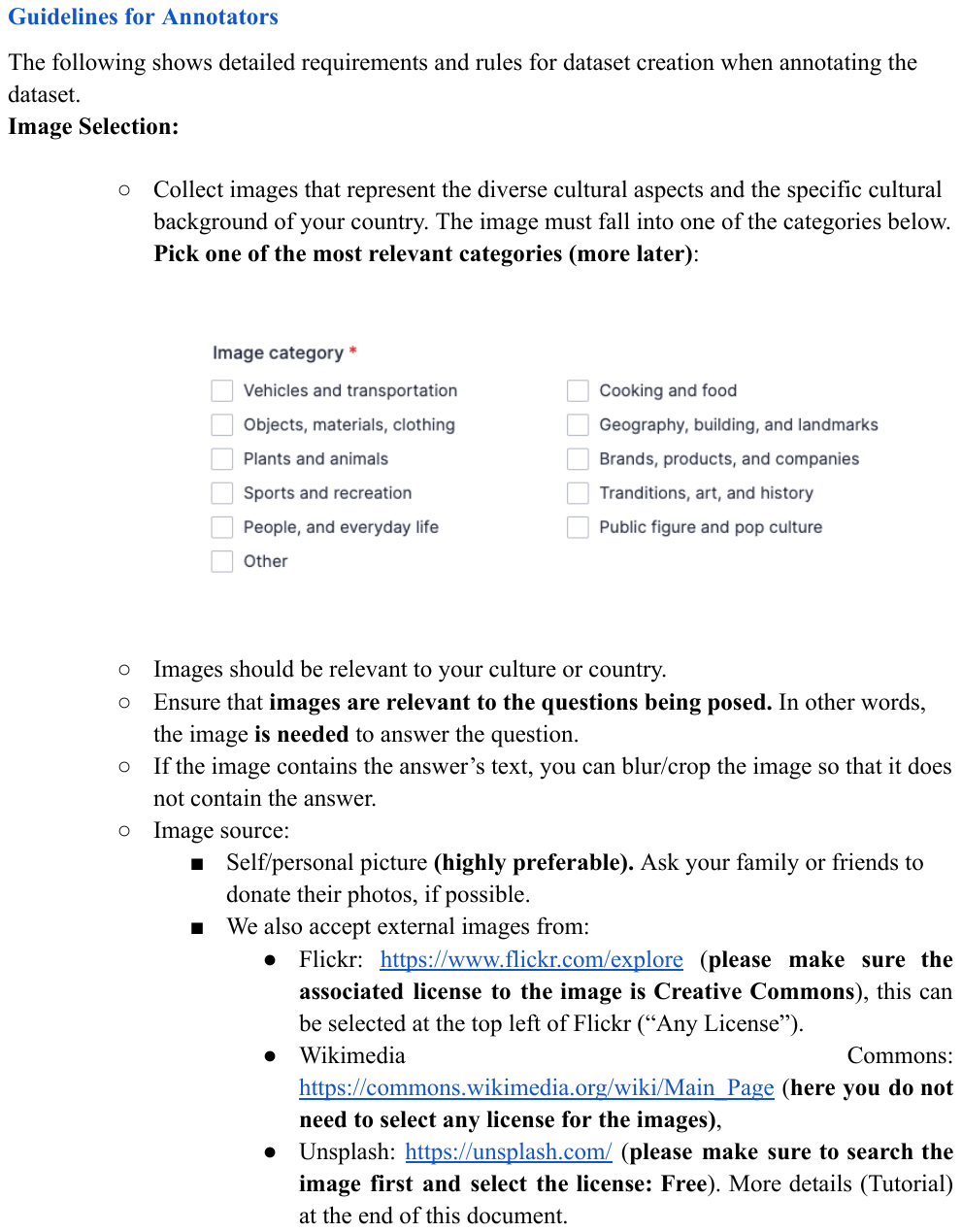}
\section{Annotator Demography}\label{demo}

\begin{table}[ht]
\centering
\resizebox{0.65\linewidth}{!}{%
\begin{tabular}{cccc}
\hline
\textbf{Annotator ID} & \textbf{Gender} & \textbf{Age Group} & \textbf{Resides in Africa} \\ \hline
Ethiopia-amh  & M & 18-30 & \checkmark \\ 
Nigeria-hau& F & 30-40 & \checkmark \\ 
Nigeria-ibo  & F & 0-40 & \checkmark \\ 
Uganda-lug  & F & 18-30 & \checkmark \\ 
Ethiopia-orm & M & 30-40 & \checkmark \\ 
Rwanda-kin  & F & 18-30 & \checkmark \\
Kenya-kik  & F & 30-40 & \checkmark \\ 
Somali-som & F & 18-30 & \checkmark \\ 
Eritrea-tir  & M & 18-30 & \checkmark \\ 
Ghana-twi  & F & 18-30 & \checkmark \\ 
Nigeria-yor & M & 30-40 & \checkmark \\ 
Botswana-tsn  & M & 30-40 & \checkmark \\ 
Malawi-nya  & F & 18-30 & \checkmark \\ 
S.Africa-zul  & F & 18-30 & \checkmark \\ 
Lesotho-sot  & F & 30-40 & \checkmark \\ \hline
\end{tabular}}
\caption{Annotator Demographics}
\label{tab:annotator_demographics}
\end{table}

% If you want to use actual checkmarks, add this to your preamble:
% \usepackage{amssymb}
% Then you can use \checkmark for checked and -- or \times for unchecked
% \FloatBarrier
% \section{ } \label{review_guideline}
% \includepdf[pages=-]{}


\section*{Supplementary material}

## K  Review Guidelines

**Afri-MC-VQA Dataset Review Guidelines**

**Overview**

This document provides comprehensive guidelines for reviewers to ensure quality control and consistency across all submissions for the Afri-MC-VQA dataset. Reviewers must carefully evaluate each submission against these criteria to ensure the integrity of the dataset.

**1. Image Review Criteria**

**1.1 Privacy and Anonymization**

 All images must protect personal privacy

- **Faces**: All private individuals' faces must be blurred (public figures exempt)
- **Personal Information**: Must be completely obscured:
  - Vehicle registration numbers
  - Phone numbers
  - Home addresses
  - Personal identification documents
  - Credit card information
- **Verification**: Use image zoom to check for any missed PII

**1.2 Cultural Relevance**

- The image must clearly represent aspects of local culture
- Should be meaningful to the specific country/region
- Must align with one of the designated cultural categories
- Reject generic images that could be from anywhere

**1.3 Image Uniqueness**

- Check for duplicate images within the same batch
- Verify against existing submissions to avoid repetition
- Similar scenes from different angles are acceptable if they enable different questions
- Flag any stock photos or overly generic content

**2. Question Quality Standards**

**2.1 Language Quality**

L   Review Guidelines

- **English Version**:
    - Must be grammatically correct
    - Natural phrasing (not awkward translations)
    - Clear and unambiguous
- **Native Language Version**:
    - Fluent and natural in the local language
    - Appropriate register and tone
    - Correct spelling and diacritics

## 2.2 Translation Accuracy

- **Semantic Equivalence**: Both versions must convey identical meaning
- **Not Word-for-Word**: Translations should be idiomatic, not literal
- **Cultural Adaptation**: Acceptable to adapt phrasing for cultural context while maintaining meaning

## 2.3 Question Independence

- **Stand-alone**: Question must be answerable from the image alone
- **No Option Dependency**: Remove questions like "Which of these is NOT..." that require seeing options
- **Single Clear Answer**: Only one option should be definitively correct
- **Test**: Cover the options - can the question still be answered?

## 2.4 Clarity and Specificity

- **No Vagueness**: Questions must be precise
- **Object Specification**: When multiple objects present, clearly indicate which one

## 2.5 Formatting

- **No Numbering**: Questions should not start with "1.", "Q:", etc.
- **Proper Punctuation**: End with a question mark
- **Capitalization**: Follow standard rules for both languages

## 3. Multiple Choice Options Review

## 3.1 Plausibility Check

- **All Options Believable**: Someone unfamiliar with the culture should find all options possible
- **No Obvious Eliminations**: Avoid options that are clearly wrong

## M    Review Guidelines

- **Context-Appropriate**: All options should fit the question's context

### 3.2 Format Consistency

- **Uniform Structure**: All options should have similar length and detail
- **Parallel Grammar**: Same grammatical structure across options

### 3.3 Prohibited Answer Types

- **Never Accept**:
  - "None of the above"
  - "No answer"
  - "Nothing"
  - "All of the above"
  - "Not applicable"

### 3.4 Objectivity Requirements

- **No Subjective Adjectives**: Avoid beautiful, ugly, nice, bad, etc.
- **Factual Descriptions Only**: Focus on observable characteristics

## 4. Audio Review Requirements

### 4.1 Transcript Accuracy

Audio must match text EXACTLY

- **Word-for-Word**: No deviations, additions, or omissions
- **Pronunciation**: Clear and accurate for the language
- **Verification Method**: Follow along with text while listening

### 4.2 Audio Structure

- **Question Format**: Natural reading with appropriate intonation
- **Option Spacing**: Brief pause (0.5-1 second) between each option
- **Answer Position**: Correct answer must ALWAYS be the first option spoken

### 4.3 Audio Quality

- **Clarity**: No background noise or distortion
- **Volume**: Consistent throughout recording
- **Speed**: Natural speaking pace (not rushed or too slow)
- **Complete**: Full question and all options included

## N    Review Guidelines

### 5. Critical Review Points

#### 5.1 Question-Option Alignment

- **Check**: Ensure question type matches option format
- **Common Issue**: Yes/No question with descriptive options
- **Verify**: Question grammar aligns with option grammar

#### 5.2 Cultural Sensitivity

- **No Stereotypes**: Avoid reinforcing negative cultural stereotypes
- **Respectful Representation**: Ensure dignified portrayal of cultural elements
- **Balanced View**: Include both traditional and modern aspects

#### 5.3 Language Quality Control

- **Spelling**: Check both languages thoroughly
- **Grammar**: Ensure proper sentence structure
- **Punctuation**: Consistent and correct usage
- **Diacritics**: Verify correct marks for languages that require them

#### 5.4 Content Diversity

- **Avoid Repetition**: Similar questions across different images
- **Varied Complexity**: Mix of simple and complex questions
- **Different Question Types**: Not all "What is..." questions
- **Topic Distribution**: Balanced across cultural categories

#### 5.5 Question Complexity

- **Avoid Trivial Questions**:
- **Require Cultural Knowledge**: Questions should test understanding, not just observation
- **Informative Value**: Each question should teach something about the culture


\FloatBarrier
% \section{Example Appendix}
% \label{sec:appendix}

% This is an appendix.

\end{document}